\let\oldyear\year
\documentclass{ieeeaccess}
\usepackage{spotcolor}

\let\year\oldyear
\usepackage{cite}

\newcommand{\headerlogoall}{}  
\newcommand{\headerlogo}{}     
\usepackage{multirow}

\usepackage{adjustbox} 
\usepackage{subcaption}
\usepackage{amsmath,amssymb,amsfonts}
\usepackage{algorithmic}
\usepackage{graphicx}
\usepackage{textcomp}
\def\BibTeX{{\rm B\kern-.05em{\sc i\kern-.025em b}\kern-.08em
    T\kern-.1667em\lower.7ex\hbox{E}\kern-.125emX}}

\newtheorem{definition}{Definition}

\usepackage{tikz}
\usetikzlibrary{backgrounds}
\pgfkeys{/donut/.cd,
inner radius/.initial=0.7cm,
inner radius=0.7cm,
outer radius/.initial=3.14cm,
outer radius=3.14cm,
text color/.initial=white,
text color=white}
\newcommand{\donutchart}[2][]{
   \pgfmathsetmacro{\totalnum}{0}
   \foreach [count=\n] \value/\colour/\name in {#2} {
     \pgfmathparse{\value+\totalnum}
     \global\let\totalnum=\pgfmathresult
     \xdef\numitems{\n}
   }

  \begin{tikzpicture}
  \pgfmathsetmacro{\wheelwidth}{\pgfkeysvalueof{/donut/outer
  radius}-\pgfkeysvalueof{/donut/inner radius}}
  \pgfmathsetmacro{\midradius}{(\pgfkeysvalueof{/donut/outer radius}
  +\pgfkeysvalueof{/donut/inner radius})/2}

  \begin{scope}[#1]

    \pgfmathsetmacro{\cumnum}{0}
    \foreach \value/\colour/\name in {#2} {
        \pgfmathsetmacro{\newcumnum}{\cumnum + \value/\totalnum*360}

        \pgfmathsetmacro{\midangle}{-(\cumnum+\newcumnum)/2}
        \begin{scope}[on background layer]
          \filldraw[draw=white,fill=\colour]
          (-\cumnum:\pgfkeysvalueof{/donut/outer radius})
          arc(-\cumnum:-(\newcumnum):\pgfkeysvalueof{/donut/outer radius}) --
          (-\newcumnum:\pgfkeysvalueof{/donut/inner radius})
          arc(-\newcumnum:-(\cumnum):\pgfkeysvalueof{/donut/inner radius}) -- cycle;
        \end{scope}
        \draw node [text=\pgfkeysvalueof{/donut/text color},
        font=\bfseries\sffamily] at
        (\midangle:{\pgfkeysvalueof{/donut/inner radius}+\wheelwidth/2}) {\name};

        \global\let\cumnum=\newcumnum
    }

  \end{scope}

  \end{tikzpicture}}

\tikzset
{
    treenode/.style = {circle, draw=black, align=center, minimum size=1cm},
    subtree/.style  = {isosceles triangle, draw=black, align=center, minimum height=0.5cm, minimum width=1cm, shape border rotate=90, anchor=north}
}

\usetikzlibrary{positioning,quotes}

\NewSpotColorSpace{PANTONE} 

\AddSpotColor{PANTONE}{PANTONE3015C}{PANTONE\SpotSpace 3015\SpotSpace C}{0 0 0 1} 

\SetPageColorSpace{PANTONE}%

\definecolor{accessblue}{cmyk}{0,0,0,1}   
\definecolor{greycolor}{cmyk}{0,0,0,1}    

\begin{document}

\title{QCSE: A Pretrained Quantum Context-Sensitive Word Embedding for Natural Language Processing}
\author{\uppercase{Charles M. Varmantchaonala}\authorrefmark{1} \IEEEmembership{Member, IEEE},  \uppercase{Niclas Götting} \authorrefmark{1}, \uppercase{Nils-Erik Schütte} \authorrefmark{1,2}, \uppercase{Jean Louis E. K. Fendji}\authorrefmark{3,4}, and \uppercase{Christopher Gies}\authorrefmark{1}}


\address[1]{Institute for Physics, Faculty V, Carl von Ossietzky University of Oldenburg, 26129 Oldenburg, Germany
} 
\address[2]{DLR, Institute for Satellite Geodesy and Inertial Sensing, Am Fallturm 9, 28359 Bremen, Germany
}
\address[3,]{Department of Computer Engineering, University Institute of Technology, University of Ngaoundere, P.O. Box 454 Ngaoundere, Cameroon} 
\address[4]{Stellenbosch Institute for Advanced Study (STIAS),
Wallenberg Research Centre at Stellenbosch University, Stellenbosch, South Africa}



\corresp{Corresponding author: Christopher Gies (e-mail: christopher.gies@uni-oldenburg.de)}

\begin{abstract}
    Quantum Natural Language Processing (QNLP) offers a novel approach to encoding and understanding the complexity of natural languages through the power of quantum computation. This paper presents a pretrained quantum context-sensitive embedding model, called QCSE, that captures context-sensitive word embeddings, leveraging the unique properties of quantum systems to learn contextual relationships in languages. The model introduces quantum native context learning, enabling the utilization of quantum computers for linguistic tasks. Central to the proposed approach are innovative context matrix computation methods, designed to create embeddings of words based on their surrounding linguistic context. Five distinct methods are proposed and tested for computing the context matrices, incorporating techniques such as exponential decay, sinusoidal modulation, phase shifts, and hash-based transformations. These methods ensure that the quantum embeddings retain context sensitivity, thereby making them suitable for downstream language tasks where the expressibility and properties of quantum systems are valuable resources. To evaluate the effectiveness of the model and the associated context matrix methods, evaluations are conducted on both a Fulani corpus, a low-resource African language, a small-size dataset, and an English corpus of slightly larger size. The results demonstrate that QCSE not only captures context sensitivity, but also leverages the expressibility of quantum systems to represent rich, context-aware language information. The use of Fulani further highlights the potential of QNLP to mitigate the problem of lack of data for this category of languages. This work underscores the power of quantum computation in natural language processing (NLP) and opens new avenues for applying QNLP to real-world linguistic challenges across various tasks and domains.
\end{abstract}


\titlepgskip=-21pt

\maketitle

\section{Introduction}
\label{sec:introduction}

Natural Language Processing (NLP) has rapidly become a cornerstone of modern technology and industry, playing a key role in applications ranging from language translation and text generation, to virtual assistants and data understanding and analysis \cite{uddin2024review, pais2022nlp, lane2025natural, tyagi2023demystifying}. These advances have been driven by powerful machine learning models that enable computers to process, understand, reason, and generate human language \cite{kalyan2021ammus, treviso2023efficient, qin2024large}. Despite this progress, NLP models still face significant challenges, especially when it comes to understanding complex context, meaning, and nuances in language\cite{khan2023exploring, hadi2023large}. Classical models are computationally expensive and still face limitations in capturing deep contextual meaning, struggling with certain subtleties in natural language, especially in complex language structures\cite{mars2022word, gao2021limitations, wang2020survey, khan2023exploring, hadi2023large, lin2023artificial, koubaa2023exploring, li2020openai, a3, a4, a5}. As the demand for more efficient and accurate NLP solutions grows, researchers are looking for new approaches to push the boundaries. Quantum computing \cite{guarasci2022quantum, widdows2024quantum, abbaszade2021application, coecke2020foundations} offers a promising frontier in this regard. 

With its ability to process information in ways that classical computers cannot, quantum computing opens up new computation possibilities. Quantum machine learning (QML) \cite{zhang2020recent, narwane2025quantum, wang2024comprehensive, olaitan2025quantum} and QNLP \cite{varmantchaonala2024quantum, coecke2020foundations, widdows2024near} explore how the principles of quantum mechanics — like superposition and entanglement — can be applied to understand language more effectively. In recent years, several pioneering works have laid the foundations for QML in NLP. Following works \cite{schuld2021machine, schuld2015introduction, wang2024comprehensive, peral2024systematic} provide a foundational understanding of QML, explaining how quantum computers could enable new learning algorithms that go beyond classical machine learning. Although QNLP is still in its infancy, its potential to surpass classical embeddings is being actively investigated, potentially leading to breakthroughs in how language is modeled and understood. Pioneering works on QNLP models, such as the quantum language model (QLM) \cite{chen2021quantum, di2022dawn} has been explored for a variety of NLP tasks, including question answering and sentiment analysis\cite{zhang2018end, zhang2018quantum, zhang2020quantum}, where superposition and entanglement-based embeddings have shown promising results in capturing dependencies between words in sentences. The rise of QNLP has sparked interest in developing particularly quantum word and sequence embedding models \cite{li2018quantum, panahi2019word2ket, nguyen2024quantum, coecke2020foundations, karamlou2022quantum}.

Quantum word embedding aims to represent words as quantum states that inherently capture their context and meaning for subsequent NLP tasks. A wide range of quantum embedding models has been proposed to handle specific language representation challenges. One such model is the complex-valued embedding model (CVE) \cite{li2018quantum, harvey2024learning, zhang2022complex, jaiswal2018quantum}, which uses quantum-inspired representations to map words onto complex-valued vectors, capturing richer information compared to real-valued embeddings. The CVE model has been applied with promising results to NLP tasks such as sentiment analysis and classification \cite{chu2024effective, zhao2022quantum, shi2021two}. Similarly, the quantum-inspired pretrained model \cite{shi2024pretrained} is designed to integrate quantum principles with classical deep neural networks, offering a hybrid approach to learning word embeddings with pretraining strategies commonly found in NLP. Other recent contributions, which exploit quantum entanglement are word2ket \cite{panahi2019word2ket} and the quantum language model with entanglement embedding (QLM-EE) \cite{chen2021quantum}. QLM-EE adapts classical pretrained architectures with quantum principles for text understanding. These models mark significant progress in the introduction of quantum-based techniques to NLP and open the door to future innovations in the field.

Despite their potential, current quantum embedding models face limitations that hinder their ability to fully exploit the advantages of quantum computing. One of the most notable limitations is their reliance on pre-trained classical embedding models, such as GloVe, RoBERTa, or word2vec\cite{chu2024effective, chen2021quantum, shi2024pretrained} or random initialization, to generate initial word vectors. These learned classical embeddings or vectors are then mapped onto quantum states, rather than training the quantum embedding models directly from the corpus to learn quantum-native and contextual relationships. This dependence could lead to the propagation of the inherent limitations of classical models through the subsequent stages of the quantum process by inheriting the constraints of classical representations.

 In this work, we design a quantum word embedding model for learning quantum embeddings directly from the corpus, without relying on pretrained classical embeddings or random initialization. By learning initial word contexts in a purely quantum-native framework, the model could better capture the polysemy and complex semantic relationships that are inherent in natural language.


The purpose of this work is to explore the potential of quantum word and sequence embedding and to discuss how it could enhance the performance and capabilities of NLP systems. The key contributions are the following.

\begin{itemize}
    \item Introduce a quantum embedding model that directly learns the context of words from the corpus, bypassing the need for pre-trained classical word embeddings\cite{chu2024effective, chen2021quantum, shi2024pretrained}. Unlike existing models that rely on classical pre-trained word vectors, the proposed quantum model directly learns from the word co-occurrence data in the corpus. This approach enables quantum states to be constructed without dependency on traditional embedding models, offering a fully quantum-native representation of words and their contextual relationships.

    \item Various methods are proposed for computing context matrices that serve as the foundation for encoding word relationships into quantum states. These techniques ensure that word proximity and dependencies are effectively captured within quantum circuits, facilitating dynamic quantum word embedding.

    \item Propose a quantum circuit with the balance between circuit depth and expressibility. The quantum circuits are optimized to efficiently encode word contexts with a manageable and constant number of qubits, ensuring scalability of the model for larger vocabularies and longer sequences.
    
    \item In the end, the paper provides an evaluation of the model's performance across different dataset sizes. The losses and precision of quantum embeddings are evaluated, illustrating the advantages of quantum-native approaches in terms of context capture and semantic precision.
\end{itemize}

The remainder of this paper is organized as follows. Section \ref{sec:4} provides an overview of NLP and QNLP, focusing on key concepts for understanding sake and transitioning to QNLP. In particular, Section \ref{sec:5} introduces the foundational principles of quantum computing concepts, essential for understanding the subsequent sections. Section \ref{sec:6} explores the background of QNLP, with a focus on quantum information encoding. Section \ref{sec:7} details the proposed quantum word embedding model, including the model overview and its different components. Section \ref{sec:9} provides a complexity analysis of the model in terms of number of qubits and quantum gates. Section \ref{sec:8} discusses the different methods for computing context matrices and vectors, which are critical to capturing and encoding semantic relationships in quantum circuits. To asses the performance of the model, Section \ref{sec:10} presents the evaluations carried out to validate our approach, followed by a discussion of the results. From the discussion, Section \ref{sec:12} outlines future directions for research in QNLP, and finally Section \ref{sec:13} concludes the paper with a summary of our findings and their implications.

\section{Preliminaries}
\label{sec:4}
\subsection{Quantum Computing Concepts}
\label{sec:5}
To provide a foundational understanding of the key concepts driving quantum computing and QNLP, this section introduces the fundamental principles of bits vs. qubits, superposition, entanglement, and quantum states with \(m\) qubits, along with their mathematical formulations \cite{nielsen2010quantum}. These concepts form the basis for exploring quantum word embeddings and their applications in NLP.

\begin{definition}
      A classical bit is the fundamental unit of classical information, which can exist in one of two discrete states, \(0\) or \(1\). A qubit is the quantum analog of a classical bit. Unlike a classical bit, a qubit can exist in a superposition of states, represented as:
    \begin{equation}
    |\psi\rangle = \alpha|0\rangle + \beta|1\rangle
    \label{qubit}
    \end{equation}
    where \(\alpha, \beta \in \mathbb{C}\) are complex numbers satisfying the normalization condition:
    \begin{equation}
    |\alpha|^2 + |\beta|^2 = 1.
    \end{equation}
    Here, \(|0\rangle\) and \(|1\rangle\) are the computational basis of the qubit.
\end{definition}

\begin{definition}
Superposition is a fundamental principle of quantum mechanics that allows a quantum system to exist in multiple states simultaneously. For a single qubit, the state is a linear combination of the basis states, as in Equation \eqref{qubit}.
\end{definition}

Thus, a quantum state with \(m\) qubits is described by a vector in a \(2^m\)-dimensional Hilbert space. The general form of an \(m\)-qubit quantum state is
\begin{equation}
|\psi\rangle = \sum_{i=0}^{2^m-1} \alpha_i |i\rangle,
\end{equation}
where \(|i\rangle\) represents the \(i\)-th computational basis state (e.g., \(|000\ldots0\rangle, |000\ldots1\rangle, \ldots, |111\ldots1\rangle\)), and \(\alpha_i\) are complex amplitudes satisfying the normalization condition:
\begin{equation}
\sum_{i=0}^{2^m-1} |\alpha_i|^2 = 1
\end{equation}

\begin{definition}
Entanglement\cite{horodecki2009quantum, guhne2009entanglement} is a quantum phenomenon in which two or more quantum systems become correlated in such a way that the state of one cannot be described independently of the others. 

The most simple relevant entangled states are the Bell states. One of these Bell states can be written as
\begin{equation}
|\psi\rangle = \frac{1}{\sqrt{2}}(|00\rangle + |11\rangle).
\end{equation}
This state cannot be factored or separated into individual qubit states, i.e., \(|\psi\rangle \notin \{|\psi_1\rangle \otimes |\psi_2\rangle: |\psi_1\rangle, |\psi_2\rangle \in \mathcal{H}\}\).
\end{definition}

\begin{definition}
Quantum gates are reversible operations that manipulate the states of qubits, enabling superposition and entanglement to process quantum information. 
Single-qubit Pauli gates act as 
\[
X|0\rangle = |1\rangle,\quad X|1\rangle = |0\rangle \quad \text{(bit flip)},
\]
\[
Z|0\rangle = |0\rangle,\quad Z|1\rangle = -|1\rangle \quad \text{(phase flip)},
\]
\[
Y|0\rangle = i|1\rangle,\quad Y|1\rangle = -i|0\rangle \quad \text{(bit and phase flip)},
\]
with matrix representations:
\begin{align}
X &= \begin{pmatrix} 0 & 1 \\ 1 & 0 \end{pmatrix}, \quad 
Y = \begin{pmatrix} 0 & -i \\ i & 0 \end{pmatrix}, \quad 
Z = \begin{pmatrix} 1 & 0 \\ 0 & -1 \end{pmatrix}.
\end{align}
On a composite Hilbert space $\mathcal{H} = \mathcal{H}_1 \otimes \mathcal{H}_2$, two-qubit gates include the controlled-NOT that flips the target qubit if the control qubit is $|1\rangle$:
\begin{equation}
\text{CNOT}_{1,2} = |0\rangle\langle 0| \otimes I + |1\rangle\langle 1| \otimes X.
\end{equation}
The $\text{CNOT}$ gate can be used to generate Bell states:
\[
\text{CNOT}_{1,2} \frac{1}{\sqrt{2}}(|0\rangle + |1\rangle) \otimes |0\rangle 
= \frac{1}{\sqrt{2}}(|00\rangle + |11\rangle).
\]
The controlled-Z gate applies a phase flip when both qubits are $|1\rangle$:
\begin{equation}
\text{CZ}_{1,2} = |0\rangle\langle 0| \otimes I + |1\rangle\langle 1| \otimes Z = \text{diag}(1, 1, 1, -1).
\end{equation}
\end{definition}

\subsection{Quantum Natural Language Processing}
\label{sec:6}

Natural Language Processing involves two key stages: word embedding and model training. Given a vocabulary $V$, words are mapped to $d$-dimensional vectors via an embedding function $f: V \rightarrow \mathbb{R}^d$. Popular methods include \textbf{word2vec} \cite{mikolov2013distributed} and \textbf{GloVe} \cite{pennington2014glove} for maximizing the log-likelihood of context words $w_c$ given center word $w$:
\begin{equation}
\mathcal{L} = \sum_{(w,w_c)\in D} \log P(w_c|w), \quad P(w_c|w) = \text{softmax}(Ww_c + b)
\end{equation}
where $W \in \mathbb{R}^{k\times d}$ and $b \in \mathbb{R}^k$ are learnable parameters.

These embeddings ${x_1,...,x_T}$ are then processed by sequential models such as RNNs/LSTMs \cite{tarwani2017survey,xiao2020research,yao2018improved} and transformers \cite{gillioz2020overview,patwardhan2023transformers} for learning patterns over sequences of words. In a recurrent model like LSTM, the hidden states $h_t$ are computed as a function of the previous hidden states and the current input: $h_t = f(h_{t-1}, x_t)$. The training minimizes for example a cross-entropy loss $\mathcal{L}_{\text{train}} = - \sum_{i=1}^N y_i \log \hat{y}_i$, where $y_i$ is the true label and $\hat{y}_i$ is the predicted probability for sample $i$.

QNLP leverages the principles of quantum computing to enhance traditional NLP tasks. A key concept in QNLP is the use of quantum word embedding, where classical data (words or sentences) are embedded into quantum states. Instead of representing words as classical high-dimensional vectors, each word is encoded as a quantum state $|\psi\rangle$ in a Hilbert space. In this approach, every word $w$ in the vocabulary $V$ is mapped to a quantum state $|\psi_w\rangle$, where the probability amplitudes of the qubits store information about the word. For example, with amplitude encoding, a classical word vector $\mathbf{v} \in \mathbb{R}^d$ can be normalized and stored in a quantum state as $|\psi\rangle = \sum_{i=1}^{d} v_i |i\rangle$. Quantum word embeddings could leverage the superposition and entanglement properties of quantum systems to capture richer contextual relationships between words than classical embeddings.

Training QNLP models typically involves a QML framework such as the variational quantum circuit (VQC)\cite{cerezo2021variational, schuld2021machine, havlivcek2019supervised}, where the circuit parameters are optimized to minimize a loss function. Let $|\psi_w\rangle$ be the initial quantum state representing a word or sentence. The VQC transforms this state into another state through a series of unitary operations, $|\psi_{\text{output}}\rangle = U(\theta) |\psi_w\rangle$, where $U(\theta)$ is a parameterized unitary matrix dependent on the parameters $\theta$. These parameters are updated via classical optimization algorithms, such as gradient descent, minimizing a loss function $\mathcal{L}(\theta)$ computed from quantum measurements. In QNLP, such training often involves measuring the expectation values of quantum operators to compute probabilities for different outcomes, similar to classical model training. In the case of the quantum embedding model, the final model aims at maximizing the likelihood $P(|w_{\text{context}}\rangle||w_{\text{center}\rangle}, \theta)$, where $|w_{\text{center}}\rangle$ and $|w_{\text{context}}\rangle$ are quantum states of the word to embed and its context words in a sentence, respectively, similar to the classical word embedding prediction tasks. This enables the model to learn contextual word embeddings in a quantum-enhanced space. Quantum encoding strategies are crucial for effectively transforming text into processable quantum states, offering potential advantages over classical methods in contextual representation.

\section{Proposed Quantum Word Embedding Model}
\label{sec:7}

\subsection{Model overview}

The aim is to map each word to a quantum state that captures its semantic meaning in the context of a given task. Suppose that we have a set of words \( W = \{ w_1, w_2, \dots, w_n \} \), and a vocabulary of size \( |V| \). In classical word embeddings, each word is mapped to a high-dimensional vector \( \mathbf{w}_i \in \mathbb{R}^d \), where \( d \) is the embedding dimension. For quantum embeddings, the quantum model instead maps words to quantum states \( |w_i\rangle \in \mathcal{H} \), where \( \mathcal{H} \) is a Hilbert space, and the quantum embedding process is represented by a unitary transformation that incorporates the word context. The process of learning quantum embeddings involves applying a VQC, typically referred to as an ansatz, to capture the relationships between words in the context of a given task. The ansatz parameters are updated iteratively using optimization methods, such as gradient-based optimization, to minimize a task-specific loss function, such as the cross-entropy. We propose a context matrix concept (see Section \ref{sec:8}) that transforms the surrounding terms of a word into a structured matrix that captures identifiers, positions, distances, and semantic–syntactic relations to construct the context-sensitive quantum embedding.

\begin{figure*}[!t]
\centering
\includegraphics[width=\textwidth]{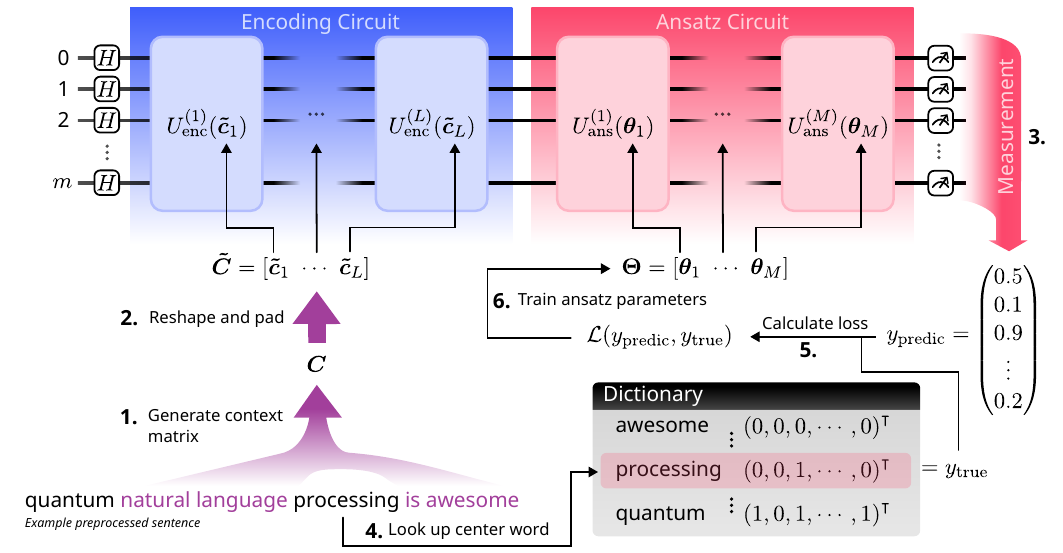}
\caption{QCSE architecture and training pipeline. The model combines: (1) a quantum context-encoding circuit, and (2) a parameterized quantum circuit trained to generate context-sensitive embeddings. Step 1: Define the context \{natural, language, is, awesome\} and identify the center word (here, processing). The context is transformed into a context matrix. Step 2: Encode the matrix using the context encoding circuit. Context matrices are reshaped for quantum encoding (no resizing required when columns match qubit count). The flexible design adapts to diverse embedding approaches. Step 3: The model predicts the quantum embedding of the center word. Steps 4–6: Compute the loss against the true embedding of the center word and optimize ansatz parameters.}
\label{full-qem}
\end{figure*}

\begin{figure}[htbp]
    \centering
    \hspace*{-5mm} 
    \includegraphics[width=\linewidth]{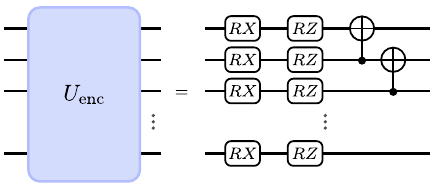}
    \hspace*{-5mm} 
    \caption{Architecture of one layer of the quantum context-encoding circuit.}
    \label{enc-circ}
\end{figure}

\begin{figure}[htbp]
    \centering
    \hspace*{-5mm} 
    \includegraphics[width=\linewidth]{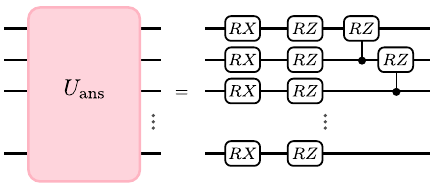}
    \hspace*{-5mm} 
    \caption{Architecture of one layer of the parameterized quantum circuit.}
    \label{ans-circ}
\end{figure}
\subsection{Model Architecture}

This section describes in detail the proposed QCSE model, which leverages parameterized quantum circuits to encode context information and uses its trainable parameters to form a VQC framework. The model consists of two main components: the context-encoding quantum circuit and the ansatz circuit with trainable parameters, as shown in Figure \ref{full-qem}. These components are combined to form the full QCSE model, which is used to predict the quantum state for the given word context data.

\subsubsection{Quantum Context Embedding Circuit}

The first component of the QCSE model is the context encoding circuit (shown in blue in Figure \ref{full-qem}), which transforms classical data (i.e., the context matrix or vector) into a quantum state. The context encoding circuit, depending on the size of the context matrix, is made up of several layers and each layer is a set of rotation gates with entanglement as shown in Figure \ref{enc-circ}. Given a context matrix $\mathbf{C}\in \mathbb{R}^{n \times n}$ (see Sec.~\ref{sec:8}), 
we reshape it into \(\mathbf{\tilde{C}} \in \mathbb{R}^{2m \times \lceil{n^2/2m}\rceil}\). If the number of elements in the last column is less than $2m$, we pad with zeros. Here, $n$ denotes the number of context features and $m$ the number of qubits. The encoding process then applies rotation gates on the qubits to embed the context data into a quantum state, which is subsequently used to learn and predict the corresponding quantum embedding.

To formally describe the encoding process, let \( \mathbf{\tilde{C}} = [\mathbf{\tilde{c}}_1, \mathbf{\tilde{c}}_2, \dots, \mathbf{\tilde{c}}_L] \), where each \( \mathbf{\tilde{c}}_l \in \mathbb{R}^{2 \times m} \) is the set of angles for the quantum gates of one layer applied to the \( m \) qubits. The encoding circuit begins by applying a Hadamard gate \( H \) to all qubits to create a superposition state, representing the initial uncertainty of the embedding (meaning) of the word. After applying Hadamard gates, the initial state of the quantum system takes the form of:
\begin{equation}
\label{H-gates}
\lvert \psi_0 \rangle = \frac{1}{\sqrt{2^m}} \sum_{x \in \{0,1\}^m} \lvert x \rangle
\end{equation}

Subsequently, each vector \( \mathbf{\tilde{c}}_l \) in the context matrix encodes the rotation angles for the quantum gates, as the \(l\)-th layer of the encoding circuit. Specifically, for each qubit \( q \in \{1, \cdots, m\}\), we apply two rotation gates \( RX_q(\mathbf{\tilde{c}}_l^{(q,1)}) \) and \( RZ_q(\mathbf{\tilde{c}}_l^{(q, 2)}) \) around the \( x \) and \( z \)-axis of the Bloch sphere, as follows:
\begin{equation}
    \begin{aligned}
    \text{$RX_q$}(\mathbf{\tilde{c}}_l^{(q,1)}) &= \exp\left(-\text{i}\frac{\mathbf{\tilde{c}}_l^{(q,1)}}{2} X_q\right) \\
&= \begin{pmatrix} \cos(\mathbf{\tilde{c}}_l^{(q,1)}/2) & -\text{i}\sin(\mathbf{\tilde{c}}_l^{(q,1)}/2) \\ -\text{i}\sin(\mathbf{\tilde{c}}_l^{(q,1)}/2) & \cos(\mathbf{\tilde{c}}_l^{(q,1)}/2) \end{pmatrix}
  \end{aligned}
\end{equation}

\begin{equation}
    \begin{aligned}
\text{$RZ_q$}(\mathbf{\tilde{c}}_l^{(q, 2)}) &= \exp\left(-\text{i}\frac{\mathbf{\tilde{c}}_l^{(q, 2)}}{2} Z_q\right) \\
&= \begin{pmatrix} \exp(-\text{i}\mathbf{\tilde{c}}_l^{(q, 2)}/2) & 0 \\ 0 & \exp(\text{i}\mathbf{\tilde{c}}_l^{(q, 2)}/2) \end{pmatrix}
    \end{aligned}
\end{equation}
where \( \mathbf{\tilde{c}}_l^{(q,1)} \text{ and }  \mathbf{\tilde{c}}_l^{(q, 2)} \) represent the two context parameters for qubit \( q \) in each encoding layer. This encoding operation is repeated for each vector \( \mathbf{\tilde{c}}_l \) in the context matrix, ensuring that the input context is fully encoded into the quantum state.

The complete rotation operations for each layer $\mathbf{\tilde{c}}_l$ of the context matrix can be described as:

\begin{equation}
\lvert \psi_{\text{enc}} \rangle = \left( \bigotimes_{q=1}^{m} RZ_q(\tilde{c}_l^{(q,1)}) RX_q(\tilde{c}_l^{(q,2)}) \right) \lvert \psi_0 \rangle
\end{equation} where \( \lvert \psi_0 \rangle \) is the initial quantum state (typically initialized to \( \lvert 0 \rangle^{\otimes m} \) and then applying $H$-gates as in Eq. \ref{H-gates}).

After the single-qubit rotations, the circuit applies a series of controlled-NOT (CNOT) gates between adjacent qubits to introduce entanglement. These CNOT gates are applied in a cascading fashion, connecting qubit \( q \) with qubit \( q+1 \) to generate the entangled quantum state


\begin{equation}
\lvert \psi_{\text{ent}} \rangle = \text{CNOT}(m-1, m) \cdots \text{CNOT}(2, 3) \cdot \text{CNOT}(1, 2) \lvert \psi_{\text{enc}} \rangle
\end{equation}

This completes the context-encoding process, leaving the quantum system in an entangled quantum state that encodes the input context information to pass throughout the trained parameterized circuit for predicting the quantum context-sensitive embedding of the word.

For illustration purposes, we have added an example in Figure \ref{full-qem}. The example illustrates the process of transitioning from a word, \textit{processing} in this case, and defining its context \textit{(natural, language, is, awesome)} within the corpus using the bag-of-word approach \cite{sivakumar2020review} to define the context of the word (another process can be used, for example, using a character-level process). Then, the context, based on the indexes, positions, frequencies, and distances of words,  is transformed into a matrix which is then encoded in a quantum state. This context-sensitive encoding passes through the ansatz for predicting the quantum word embedding, within its context.


\subsubsection{Ansatz Circuit and Trainable Parameters}

The second component of the QCSE model is the ansatz circuit, which comprises trainable parameters. The architecture of the ansatz circuit is shown in Figure \ref{ans-circ} in the red block. Let \( \mathbf{\Theta} = [\boldsymbol{\theta}_1, \boldsymbol{\theta}_2, \dots, \boldsymbol{\theta}_M] \) represent the set of trainable parameters for the ansatz circuit containing $M$ layers. Each layer is composed of unitary operations that depend on the parameters \( \boldsymbol{\theta}_a \). Each layer parameter set \(\boldsymbol{\theta}_{a \in \{1 , \cdots ,\text{ } M \}} \) includes two subsets of trainable parameters, \( \boldsymbol{\theta}_{a_1} \in \mathbb{R}^{2 \times m} \) and \( \boldsymbol{\theta}_{a_2} \in \mathbb{R}^{m-1} \), where \( m \) is the number of qubits. 

Each ansatz circuit layer consists of two parts:
A set of single-qubit rotations, parameterized by \( \boldsymbol{\theta}_{a_1} \) and a set of controlled-RZ (CRZ) gates, parameterized by \( \boldsymbol{\theta}_{a_2} \). For each qubit \( q \), the ansatz circuit applies the following operations:

\begin{equation}
\text{$RX_q$}(\theta_{a_1}^{(q,1)}) = \exp\left(-\text{i}\frac{\theta_{a_1}^{(q,1)}}{2} X_q\right),
\end{equation}
\begin{equation}
\text{$RZ_q$}(\theta_{a_1}^{(q, 2)}) = \exp\left(-\text{i}\frac{\theta_{a_1}^{(q, 2)}}{2} Z_q\right)
\end{equation}
where \( \theta_{a_1}^{(q,1)} \) and \( \theta_{a_1}^{(q, 2)} \) are the two parameters from \( \boldsymbol{\mathbf{\theta}_{a_1}} \) for the qubit  \( q \).

Next, the ansatz circuit introduces entanglement using controlled-RZ (CRZ) gates between consecutive qubits, defined as
\begin{equation}
\text{CRZ}_{q,q+1}(\phi_{q,q+1}) = \begin{pmatrix} 
1 & 0 & 0 & 0 \\ 
0 & 1 & 0 & 0 \\ 
0 & 0 & e^{-\text{i}\phi_{q,q+1}/2} & 0 \\ 
0 & 0 & 0 & e^{\text{i}\phi_{q,q+1}/2} 
\end{pmatrix}
\end{equation}
where \( \phi_{q,q+1} \) is the rotation angle parameter from \( \boldsymbol{\theta}_{a_2} \) applied between qubits \( q \) and \( q+1 \).

The complete ansatz operation for the \( a \)-th layer can be written as:

\begin{equation}
\begin{split}
\lvert \psi_{\text{ansatz}} \rangle = &\text{CRZ}_{m-1}(\phi_{m-1,m}) \cdots \text{CRZ}_1(\phi_{1,2}) \\ & \cdot \left( \bigotimes_{q=1}^{m} \text{RZ}_q(\theta_{a_1}^{(q,2)}) \text{RX}_q(\theta_{a_1}^{(q,1)}) \right) \lvert \psi_{\text{ent}} \rangle
\end{split}
\end{equation}

This completes the ansatz block, where the parameters \( \mathbf{\Theta} \) are trainable and can be optimized to minimize the loss function during the training process.

\subsection{Combination of the Components}

The complete QCSE model combines the context encoding circuit and the ansatz circuit into a single quantum circuit. This quantum circuit is used to perform the final measurement, which predicts the qubit string probabilities for the input data.

After applying both the context-encoding circuit and the ansatz circuit, the expectation value of the Pauli-Z operator is measured on each qubit. For each qubit \( q \), the expectation value of the Pauli-\(Z\) operator is
\begin{equation}
\langle Z_q \rangle = \langle \psi_{\text{ansatz}} \rvert Z_q \lvert \psi_{\text{ansatz}} \rangle.
\end{equation}
This quantity takes values in the interval \([-1,1]\).
The measurement probabilities of \( q \), in the computational basis states, are
\begin{equation}
P(\lvert 0 \rangle_q) = \frac{1 + \langle Z_q \rangle}{2}, 
\quad 
P(\lvert 1 \rangle_q) = \frac{1 - \langle Z_q \rangle}{2}.
\end{equation}
Finally, the output of the quantum embedding model is the list of probabilities \( P(\lvert 1 \rangle_q) \) (or equivalently \( P(\lvert 0 \rangle_q) \)) for each qubit \( q \), which represent the model's prediction for the encoded context.

The quantum embedding model described in this section provides a novel method for encoding classical data into quantum states using parameterized quantum circuits. The combination of context encoding and trainable ansatz layers allows for a flexible and expressive quantum model that can be optimized to minimize a given loss function, such as the cross-entropy. The output of the model is a set of bitstring probabilities, which can be used in various quantum machine learning applications. Here, we consider it in the context of QNLP. Figure \ref{full-qem} visualizes the complete model and the whole training process. The process establishes a basic context by leveraging nearby words or characters to generate a context matrix that encodes the quantum state of the word for further processing.

\subsection{Complexity Analysis of the Model}
\label{sec:9}

In QCSE, the number of qubits required for the quantum encoding process is driven primarily by the size of the vocabulary, not by the context size. This is because the context matrix, rather than increasing the number of qubits, is flattened, reshaped and padded with zeros when required, and encoded sequentially across multiple layers of the quantum circuit. The same set of qubits is reused for the encoding, ensuring that the qubit requirement remains constant.

The number of qubits $m$ required by the model is governed by the size of the vocabulary $|V|$. If each word in the vocabulary is represented by a computational basis state, the number of qubits is proportional to the logarithm of the vocabulary size, 
\[
m = \lceil \log_2 |V| \rceil.
\]
For a large vocabulary of size $|V| \sim 10^5$, such as the english vocabulary, this results in $m \sim 17$ qubits.

While the number of qubits remains constant, the quantum gate complexity increases with the size of the context window and the desired depth of the ansatz layers. The complexity of the quantum gate operations can be broken down into two primary components: (1) the context encoding block  and (2) the ansatz block.

\begin{itemize}
\item Context Encoding Block Complexity: Each layer corresponds to one portion of the context matrix being mapped onto the qubit system. Since the number of qubits is fixed, the encoding of the context matrix is distributed across multiple layers. Therefore, the complexity of the gates of the context encoding block scales linearly with the number of context encoding layers $L$. The total number of gates for the context encoding block is

\begin{equation}
G_{\text{context}} = (3m - 1) \cdot L.
\end{equation}

\item Ansatz Block Complexity: The ansatz block is repeated $n_{\text{repeats}}$ times. For a system of $m$ qubits, each layer of the ansatz consists of $2m$ single-qubit rotation gates and $m-1$ controlled entangling gates (such as $\text{CNOT}$ or $\text{CRZ}$). Thus, the number of gates in a single ansatz layer is $G_{\text{layer}} = 3m - 1$. The total number of gates required for the ansatz block with $n_{\text{repeats}}$ repetitions is
\begin{equation}
G_{\text{ansatz}} = (3m - 1)\cdot M.
\end{equation}

\end{itemize}

The total gate complexity of the entire model is the sum of the gates required for the ansatz block and the context encoding block, i.e,
\begin{equation}
G_{\text{total}} = G_{\text{ansatz}} + G_{\text{context}} = (3m - 1)( M  + L).
\end{equation}

\section{Context Encoding: Exponential Decay with Sinusoidal Method}
\label{sec:8}

The choice of context representation is fundamental to the quantum embedding framework. Unlike classical word embedding models that typically employ flat vector representations for context words, we propose encoding the context as a structured matrix to capture richer relational information between words within the context window. This design is motivated by several considerations: (1) matrix representations naturally encode pairwise relationships, enabling the model to capture not just individual word identities but also their interactions; (2) the structured format provides multiple dimensions for encoding positional, semantic, and distance information simultaneously; (3) this approach draws inspiration from positional encoding mechanisms in transformer architectures~\cite{vaswani2017attention}, which have proven effective at capturing word order and relative positions.

Among the various encoding strategies we explored, the exponential decay with sinusoidal encoding method demonstrated consistent and robust performance across different model depths. This method defines the context matrix $\mathbf{C} \in \mathbb{R}^{n \times n}$ for a context window of $n$ words, where each element $c_{ij}$ capturing the relationship between words at positions $i$ and $j$ is computed as:

\begin{equation}
c_{ij} = e^{-\alpha |i-j|} \sin(\omega \theta_i) \cos(\omega \theta_j) + \theta_i
\label{eq:exp_decay}
\end{equation}

where $\theta_i = \text{idx}_i \cdot \frac{2\pi}{|V|}$ represents the normalized angle for the word at position $i$, with $\text{idx}_i$ denoting its vocabulary index and $|V|$ the vocabulary size. The parameters $\alpha > 0$ and $\omega > 0$ control the exponential decay rate based on the word index, distance between words, and the changing position of words.

This formulation integrates three complementary encoding mechanisms. First, the exponential decay term $e^{-\alpha |i-j|}$ captures positional proximity: words closer in the context window receive stronger weighting, reflecting the linguistic intuition that nearby words typically have strong relationships. Second, the sinusoidal encoding $\sin(\omega \theta_i) \cos(\omega \theta_j)$ provides distinct representations for different vocabulary items through periodic functions, similar to positional encodings in transformers~\cite{vaswani2017attention, shaw2018self}. Third, the additive term $\theta_i$ ensures that each word maintains a unique signature based on its vocabulary position, preventing representational collapse even when words appear at identical relative positions in different contexts.

After construction, the context matrix is reshaped to $\mathbf{\tilde{C}} \in \mathbb{R}^{2m \times \lceil n^2/2m \rceil}$ and padded with zeros as needed to match the circuit architecture, where $m$ denotes the number of qubits. This reshaping distributes the context information across multiple encoding layers, allowing the quantum circuit to process the full relational structure through sequential gate operations.

While this exponential decay method proved most effective in our experiments, the design space for context encoding remains explorable. We investigated four alternative encoding strategies—index-based diagonal encoding, positional phase shift encoding, hash-based modulation, and positional angular shift—each offering different trade-offs in terms of representational capacity, computational efficiency, and numerical properties. These alternative methods and their comparative performance are detailed in the Appendix.

\section{Evaluation and Results}
\label{sec:10}

\subsection{Setup}

We evaluated the proposed QCSE model against classical word embedding baselines using two datasets: an English corpus (110 sentences, 550 tokens, 80-word vocabulary, requiring 7 qubits) and a Fulani\cite{10.11648} corpus (20 sentences, 120 tokens, 26-word vocabulary, requiring 5 qubits). The primary objective is to assess whether the quantum model can effectively capture contextual relationships between center–context word pairs under limited model capacity.

Due to differences in output encoding schemes and their corresponding loss functions, distinct hyperparameter configurations were employed for quantum and classical models. Quantum models used a learning rate of 0.0003 and L2 regularization ($\lambda = 0.001$), minimizing binary cross-entropy loss to accommodate bitwise output encoding. Classical baselines (CBOW and Skip-gram variants) employed categorical cross-entropy loss with learning rate 0.03 and L2 regularization ($\lambda = 0.0001$), reflecting their probabilistic multi-class formulation. All experiments used an 80-20 train-test split, a context window size of 4, and were trained for 50 epochs.

The accuracy was calculated as follows. For quantum models, predicted probabilities were first converted into binary bitstrings using a threshold of 0.5. The similarity score between predicted and target bitstrings was then calculated as the proportion of matching bits, and a prediction was considered correct if this similarity score was at least 0.5. In the same way, for classical models, the accuracy was defined as the percentage of context words whose predicted representations achieved a similarity score of at least 0.5 with the target representations.

\subsection{Discussion}
\label{sec:discussion}

Table~\ref{tab:results_english} presents the accuracy results for both quantum and classical models on the English dataset. The quantum models, implemented using the exponential decay sinusoidal encoding method, scale linearly with depth at 21 parameters per layer (ranging from 21 to 168 parameters for 1-8 layers). Classical baselines range from 8,000 to 48,000 parameters depending on embedding dimension.

\begin{table}[h]
\centering
\caption{Accuracy comparison on English dataset—QCSE (exponential decay method) vs. classical models}
\begin{tabular}{lcc}
\hline
\textbf{Model} & \textbf{Parameters} & \textbf{Accuracy (\%)} \\
\hline
\multicolumn{3}{c}{\textit{Classical Models}} \\
\hline
CBOW-50 & 8,000 & 65.5 \\
\textbf{CBOW-80} & 12,800 & \textbf{70.0} \\
CBOW-100 & 16,000 & 68.1 \\
CBOW-200 & 32,000 & 64.6 \\
CBOW-300 & 48,000 & 63.7 \\
SkipGram-50 & 8,000 & 53.8 \\
SkipGram-80 & 12,800 & 62.1 \\
SkipGram-100 & 16,000 & 58.8 \\
SkipGram-200 & 32,000 & 58.6 \\
SkipGram-300 & 48,000 & 58.2 \\
\hline
\multicolumn{3}{c}{\textit{Quantum Models (Exp. Decay Sinusoidal)}} \\
\hline
QCSE-1L & 21 & 50.4 \\
QCSE-2L & 42 & 64.6 \\
QCSE-3L & 63 & 67.3 \\
QCSE-4L & 84 & 70.8 \\
QCSE-5L & 105 & 67.3 \\
QCSE-6L & 126 & 66.4 \\
QCSE-7L & 147 & 84.1 \\
\textbf{QCSE-8L} & 168 & \textbf{85.8} \\
\hline
\end{tabular}
\label{tab:results_english}
\end{table}


Table~\ref{tab:results_Fulani} shows results on the smaller Fulani dataset, which provides insight into model behavior under more limited data conditions.

\begin{table}[h]
\centering
\caption{Accuracy comparison on Fulani dataset—QCSE (exponential decay method) vs. classical models}
\begin{tabular}{lcc}
\hline
\textbf{Model} & \textbf{Parameters} & \textbf{Accuracy (\%)} \\
\hline
\multicolumn{3}{c}{\textit{Classical Models}} \\
\hline
\textbf{CBOW-50} & 2,600 & \textbf{76.9} \\
CBOW-80 & 4,160 & 61.5 \\
CBOW-100 & 5,200 & 57.7 \\
CBOW-200 & 10,400 & 53.8 \\
CBOW-300 & 15,600 & 61.5 \\
SkipGram-50 & 2,600 & 47.1 \\
SkipGram-80 & 4,160 & 46.5 \\
SkipGram-100 & 5,200 & 51.3 \\
SkipGram-200 & 10,400 & 55.1 \\
SkipGram-300 & 15,600 & 54.2 \\
\hline
\multicolumn{3}{c}{\textit{Quantum Models (Exp. Decay Sinusoidal)}} \\
\hline
QCSE-1L & 15 & 46.2 \\
QCSE-2L & 30 & 34.7 \\
QCSE-3L & 45 & 46.2 \\
QCSE-4L & 60 & 69.2 \\
QCSE-5L & 75 & 77.0 \\
QCSE-6L & 90 & 73.1 \\
\textbf{QCSE-7L} & 105 & \textbf{80.8} \\
QCSE-8L & 120 & 61.5 \\
\hline
\end{tabular}
\label{tab:results_Fulani}
\end{table}



The quantum model using exponential decay sinusoidal encoding method demonstrates a clear progression in model capability with increasing depth. On the English dataset (Table~\ref{tab:results_english}), single-layer models achieve only 50.4\% accuracy indicating insufficient capacity to capture meaningful contextual relationships. Performance improves at 2-4 layers (64.6-70.8\%), with the 4-layer configuration matching the best classical baseline (CBOW-80 at 70.0\%) while using only 84 parameters compared to 12,800. In deeper configurations: the 7-layer model achieves 84.1\% accuracy, and the 8-layer model reaches 85.8\%, representing a 15.8 percentage point improvement over the best classical baseline while using 76× fewer parameters (168 vs. 12,800). This improvement at 7-8 layers suggests that the quantum model benefits from increased expressivity when sufficient circuit depth is available to exploit the relational structure encoded in the context matrices. However, the 5-layer (67.3\%) and 6-layer (66.4\%) models perform slightly worse than the 4-layer configuration. There may be several reasons for this phenomenon. One possibility is that this intermediate fluctuation may reflect an optimization instability. As circuit depth increases, the optimization landscape of some layers becomes more complex, which can lead to suboptimal local minima. Performance recovers at greater depths, suggesting that once the model surpasses the transitional region, the capacity becomes effectively utilized. Another possibility is that this phenomenon could be implicitly influenced by the dual thresholding mechanisms in the accuracy computation, in the same way that this inconsistency was observed with loss and accuracy. First, predicted probabilities are converted to bits using a 0.5 threshold; a probability shifting from 0.50 to 0.49 across epochs changes the bit assignment despite maintained model confidence, affecting similarity scores. Second, predictions are counted as correct only when the final similarity score exceeds 0.5. These thresholds can cause temporary accuracy decreases even as the model continues learning, as evidenced by subsequent accuracy recovery in the training curves. 

\begin{figure}[t]
\centering
\includegraphics[width=\linewidth]{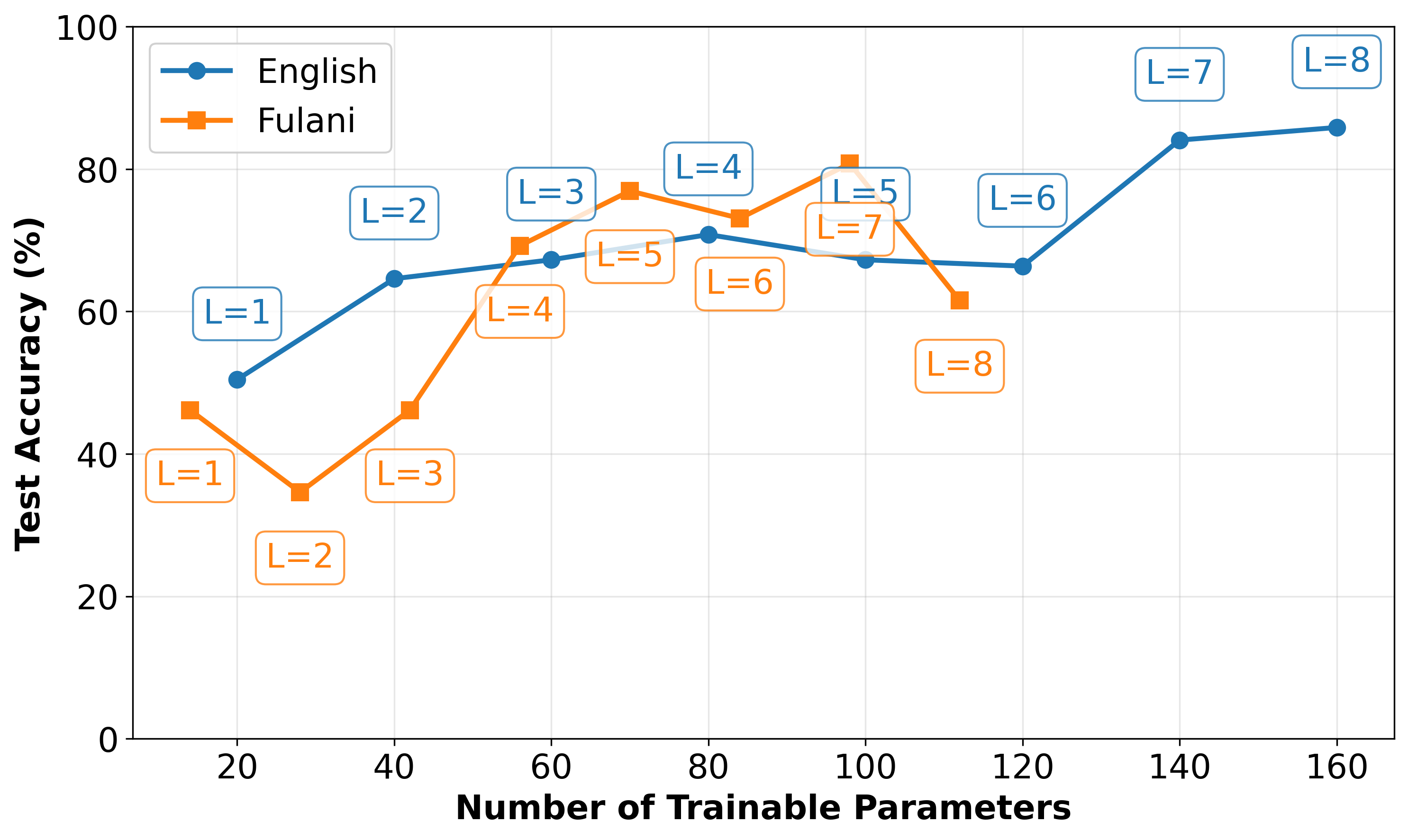}
\caption{Accuracy versus trainable parameter count in QCSE using Exponential Decay Sinusoidal Method on English and Fulani datasets.}
\label{fig:expo_acc_params}
\end{figure}

On the Fulani dataset (Table~\ref{tab:results_Fulani}), the optimal depth shifts to 7 layers (80.8\%), with the 8-layer model exhibiting degraded performance (61.5\%). This suggests dataset-dependent capacity requirements: smaller datasets may not provide sufficient training information to effectively optimize deep circuits, leading to capacity saturation or optimization difficulties beyond a certain model depth. In Figure~\ref{fig:expo_acc_params}, for the English corpus, performance increases to a maximum at 168 parameters (85.8\%) and for the smaller Fulani corpus, the peak occurs earlier at 105 parameters (80.8\%), followed by a decline. This shift in optimal parameter count (168 → 105) reflects adaptation to dataset size. Quantum models remain parameter-efficient, but require alignment between circuit depth and dataset size to achieve optimal performance. This degradation resembles the capacity–generalization trade-off observed in classical models~\cite{mikolov2013efficient}. Classical models exhibit a known phenomenon in classical embedding ~\cite{mikolov2013efficient}: when all other parameters remain constant, increasing the embedding dimension tends to degrade performance. On the English dataset, CBOW achieves comparable accuracy at 80 dimensions (70.0\%), with performance declining at higher dimensions: 68.1\% (100-dim), 64.6\% (200-dim), and 63.7\% (300-dim). Skip-gram shows a similar pattern, with performance dropping from 62.1\% (80-dim) to 58.2\% (300-dim). This behavior reflects a fundamental challenge in classical embedding models: higher-dimensional representations increase model capacity, but without sufficient training data, they become increasingly difficult to optimize and prone to overfitting. On the moderate-sized English corpus (550 tokens), models with 200-300 dimensional embeddings have too many parameters (32,000-48,000) to effectively constrain through the available training samples, resulting in poor generalization despite lower training loss. The Fulani dataset follows the same effect: CBOW-50 achieves 76.9\% accuracy with 2,600 parameters, but performance collapses at higher dimensions (53.8\% for 200-dim, 61.5\% for 300-dim). This demonstrates that classical embedding models, designed and optimized for large-scale corpora with millions of training examples, struggle to adapt their capacity to small-dataset regimes. The quantum model exhibits similar challenges. For example, on the Fulani dataset, the 8-layer quantum model drops to 61.5\% despite the 7-layer model achieving 80.8\%. This suggests that with fewer parameters (120 vs. 15,600 for CBOW-300), quantum circuits could still exceed optimal capacity for very small datasets.

Analysis of training dynamics indicates that deeper circuits (7 to 8 layers) start with higher initial loss but exhibit slightly sustained descent, achieving higher accuracy. Some shallow models converge quickly with lower final loss, but yield poor accuracy. The corresponding analysis of loss trajectories, convergence regimes, and comparisons together with all alternative encoding methods is provided in the Appendix~\ref{app:alternative_encodings}.

\section{Applications and Future Directions}
\label{sec:12}

The proposed quantum embedding model enables key NLP applications , such as language modeling, translation, and text generation by capturing linguistic relationships. QCSE can be extended to quantum encoder-decoder frameworks \cite{chen2020quantum, bausch2021quantum, zhao2024qksan, shi2024qsan} for tasks like translation and summarization. Quantum states encode context-sensitive embeddings, while decoders reconstruct classical outputs, leveraging quantum parallelism for efficiency. Quantum self-attention \cite{zhao2024qksan, shi2024qsan} could compute superposition-based attention scores, reducing the quadratic complexity of classical transformers \cite{vaswani2017attention}. Integrating quantum embeddings into pretrained models (e.g., BERT, GPT \cite{devlin2018bert}) may enhance context sensitivity. Using the right number of trainable parameters, QCSE could benefit low-resource languages \cite{magueresse2020low, joshi2019unsung} (e.g., Fulani) by extracting richer semantics from limited data.

Key directions include:

\begin{itemize}
    \item Impact of Circuit Depth: Increasing the number of layers in the word-context encoding circuit or the ansatz could enhance the model’s capacity to capture more complex semantic structures. A systematic study of different circuit depths would help identify the trade-offs between circuit expressiveness and noise-induced errors.
    \item Different Ansatz Exploration: Different types of parameterized quantum circuits~\cite{sim2019expressibility} should be explored. A more suitable ansatz could potentially improve the encoding of intricate word relationships, especially in low-resource settings.
    \item Train the QCSE model on large-scale datasets using varied training parameters to evaluate its scalability, performance, and robustness across different configurations.
\end{itemize}

\section{Conclusion}
\label{sec:13}

This work presented a pretrained Quantum Context-Sensitive Embedding (QCSE) model that leverages quantum computing principles to address NLP challenges. The study demonstrated the feasibility of applying quantum circuits to word embedding tasks, showing potential for capturing semantic relationships. The approach encodes word–context relationships as quantum states, exploiting quantum properties such as superposition and entanglement to represent linguistic patterns. Our results show that the proposed QCSE model can capture contextual relationships of texts, even if the provided corpus is limited. Overall, this work contributes to the foundation for further exploration of quantum word embeddings, particularly for low-resource NLP applications.

\section*{Acknowledgements}

Charles M. Varmantchaonala is grateful for financial support via a research grant from the German Academic Exchange Service (DAAD).

\bibliographystyle{IEEEtran}
\bibliography{mybibfile.bib}

\begin{IEEEbiography}[{\includegraphics[width=1in,height=1.35in,clip,keepaspectratio]{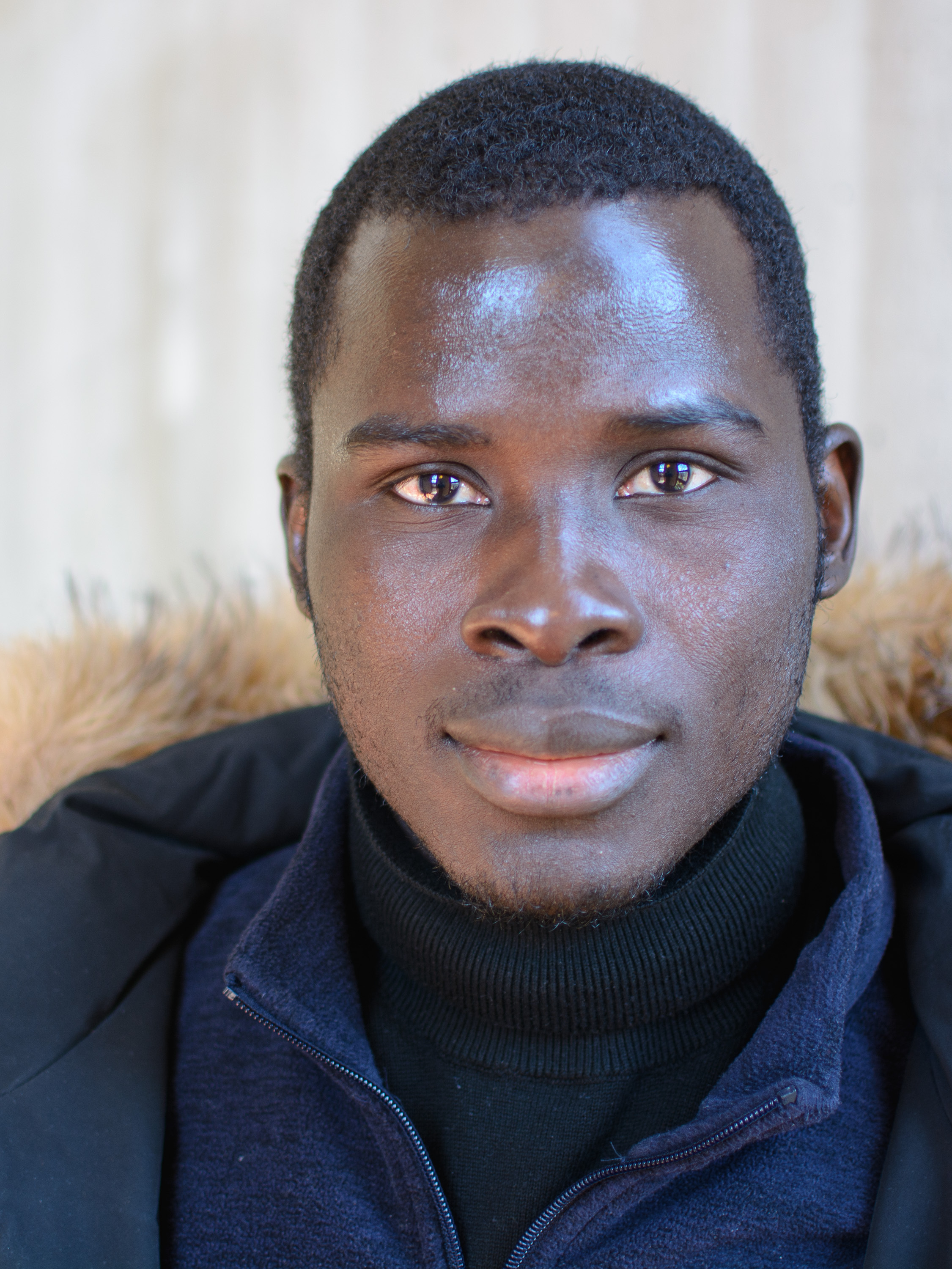}}]{Charles M. Varmantchaonala} received his B.S. degree in Mathematics and Computer Science from the University of Ngaoundere, Cameroon in 2019 and the M.Sc. degree in Systems and Software in Distributed Environments from the University of Ngaoundere in 2021. He is currently working towards a doctoral degree in physics at the Carl von Ossietzky University Oldenburg, specializing
in Natural Language Processing and Quantum
Computing. His research interests include quantum natural language processing, quantum machine learning, machine learning and deep learning, data science, software engineering, autonomous systems, and algorithms optimization.
\end{IEEEbiography}

\begin{IEEEbiography}[{\includegraphics[width=1in,height=1.25in,clip,keepaspectratio]{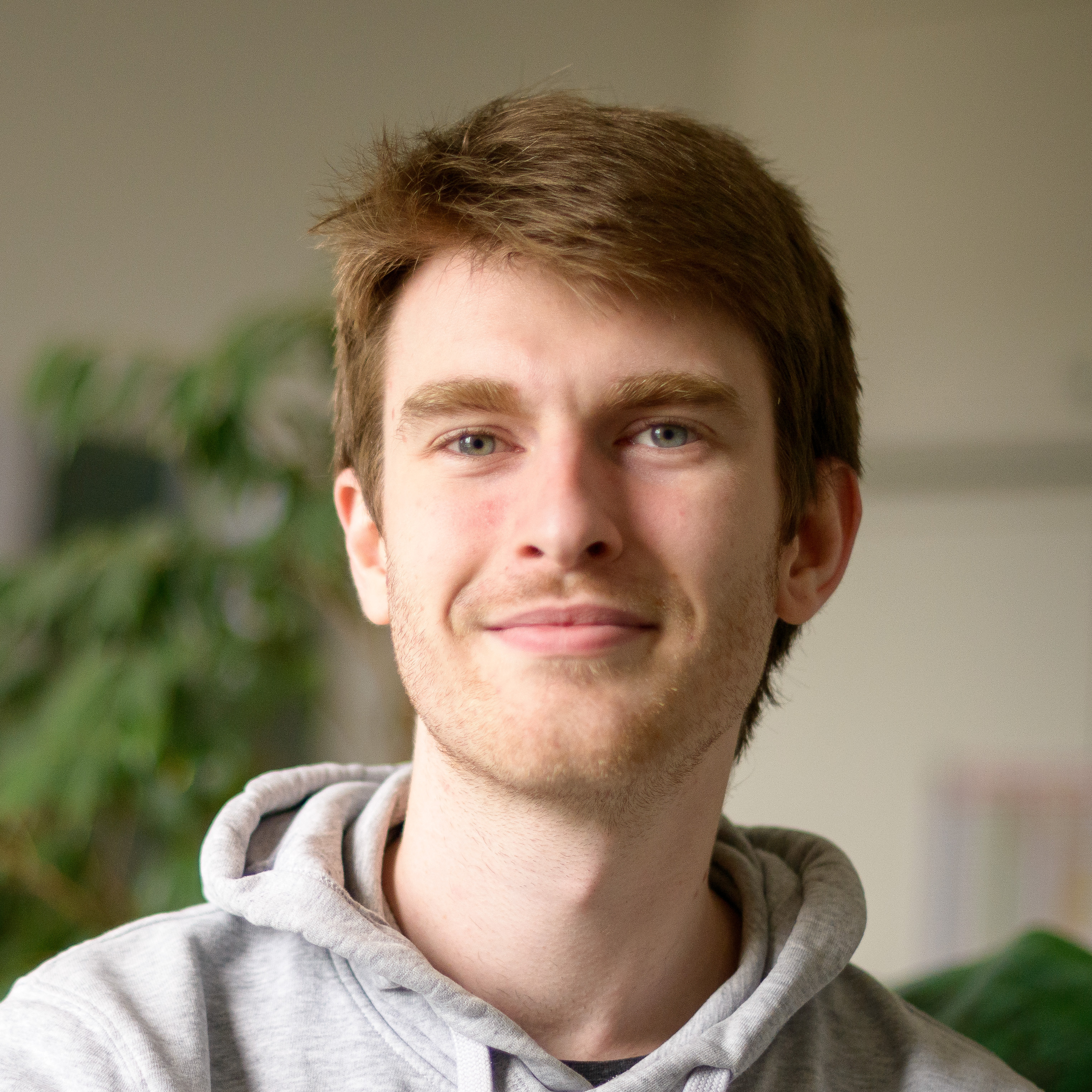}}]{NICLAS GÖTTING} earned his M.Sc. in Physics from the University of Bremen, Germany, where he focused on semiconductor physics and, in particular, on correlated Moiré physics in two-dimensional materials. He is currently a Ph.D. candidate at the Carl von Ossietzky University of Oldenburg, working on Quantum Machine Learning with a special emphasis on Quantum Reservoir Computing in open quantum systems. His scientific interests include machine learning for quantum systems, low‑dimensional condensed matter physics, quantum natural language processing, and the study of correlated phenomena. 
\end{IEEEbiography}

\begin{IEEEbiography}[{\includegraphics[width=1in,height=1.25in,clip,keepaspectratio]{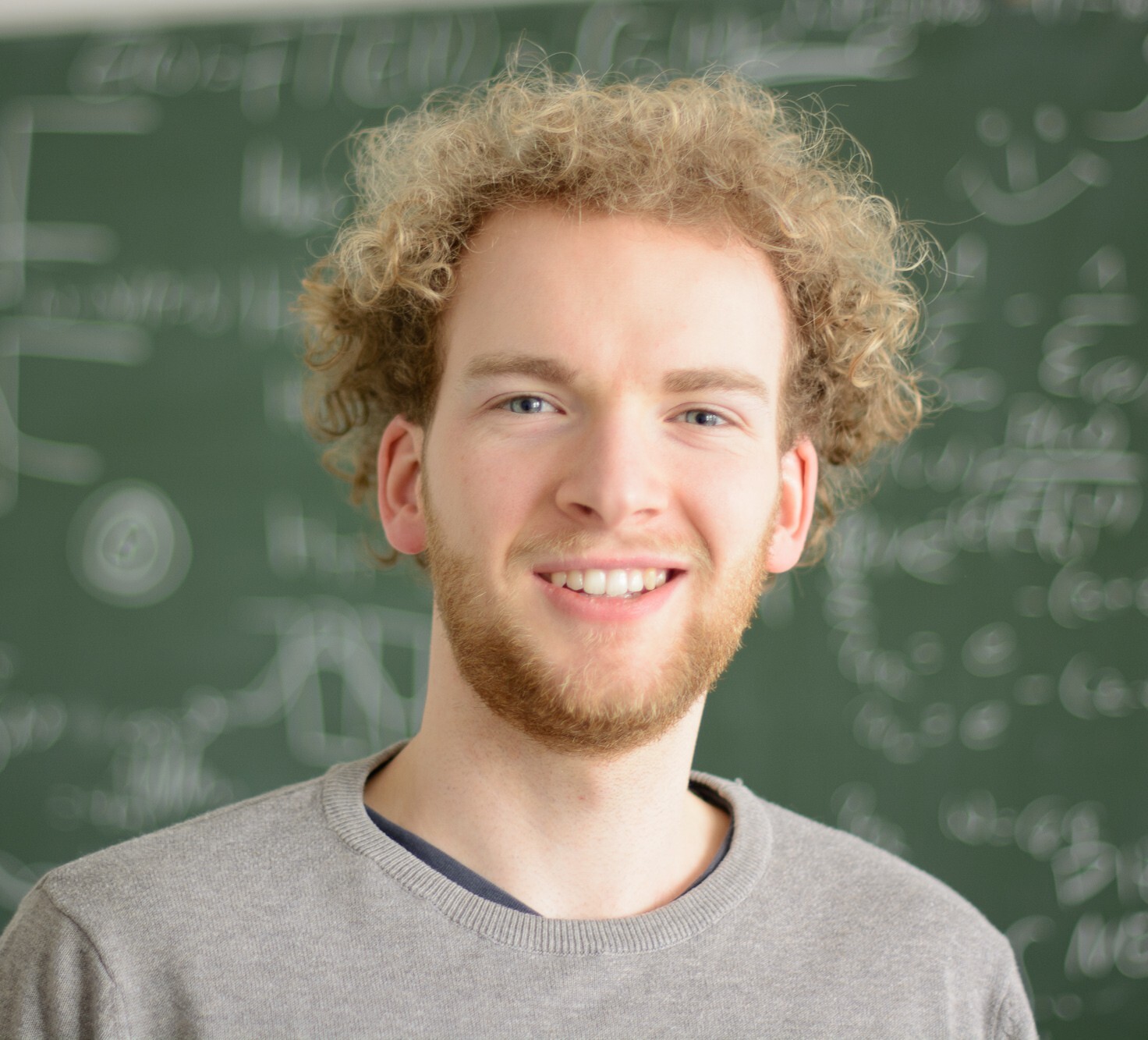}}]{NILS-ERIK SCHÜTTE} studied physics at the University of Bremen. Under the supervision of Prof. Dr. Christopher Gies, he studied the correlated behavior of interlayer excitons in moiré heterostructures  and finishing his Master’s degree in 2023. Currently, he is a research associate and Ph.D. candidate at the German Aerospace Center (DLR) in the Institute for Satellite Geodesy and Inertial Sensing lead by  Prof. Dr. Meike List. The project is a cooperation with the group of Prof. Dr. Christopher Gies at the University of Oldenburg. His current research focusses on quantum machine learning, where he investigates two paradigms, namely gate-based quantum computing and quantum reservoir computing.
\end{IEEEbiography}

\begin{IEEEbiography}[{\includegraphics[width=1in,height=1.25in,clip,keepaspectratio]{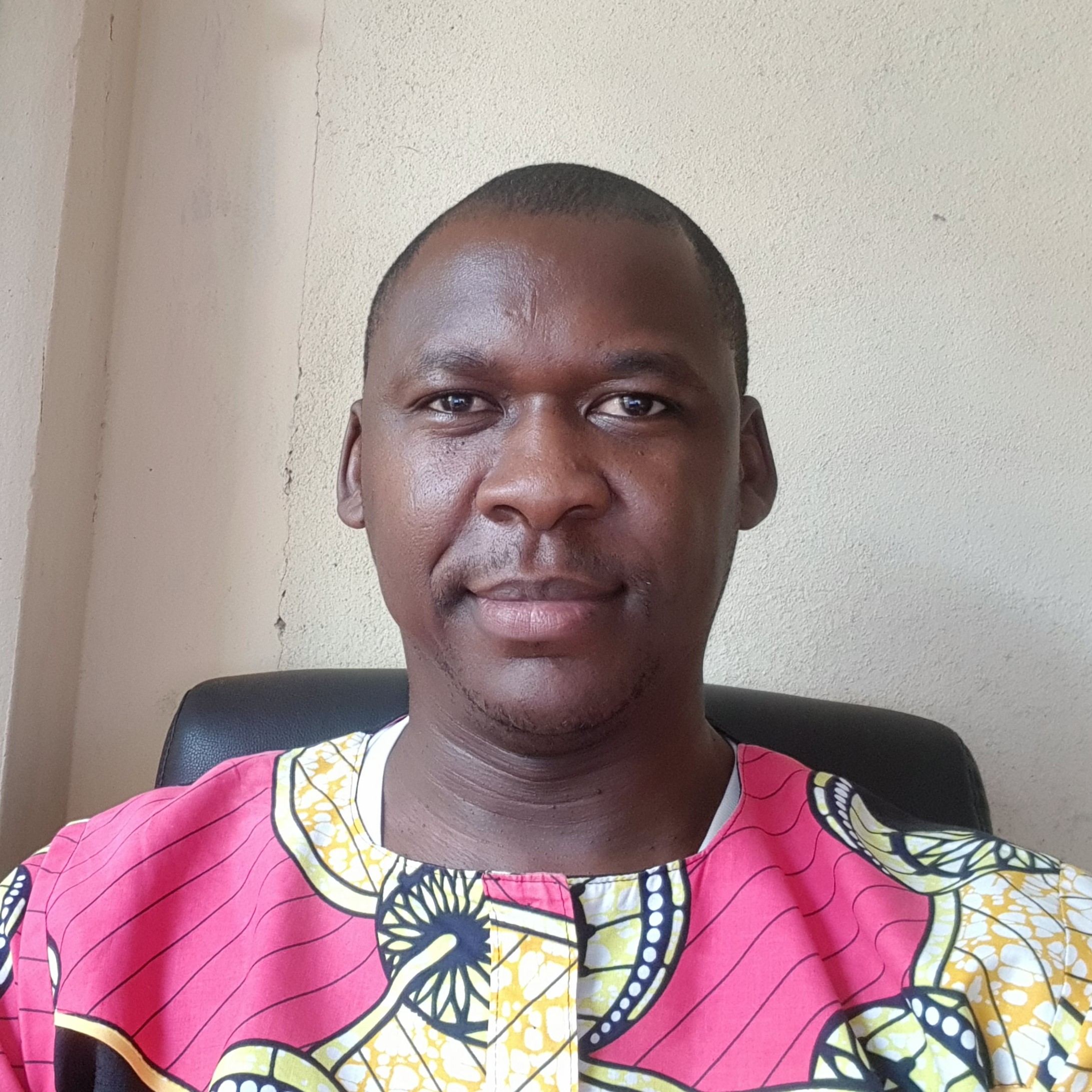}}]{Jean Louis Fendji Kedieng Ebongue} is an Associate Professor of Computer Science at the University of Ngaoundere, Cameroon and a member of the ICT and Artificial Intelligence Commission at the Ministry of Scientific Research and Innovation. He holds a Doctor of Engineering (Dr.-Ing.) in Computer Science from the University of Bremen, Germany. His research interests include Artificial Intelligence, Natural Language Processing, digitization in agriculture, data justice, ICT for development. He is a fellow at both the Hamburg Institute for Advanced Study (HIAS) and the Stellenbosch Institute for Advanced Study (STIAS). Fendji is also the Head of a Centre for Research, Experimentation and Production at the University of Ngaoundere.
\end{IEEEbiography}
\EOD

\begin{IEEEbiography}[{\includegraphics[width=1in,height=1.25in,clip,keepaspectratio]{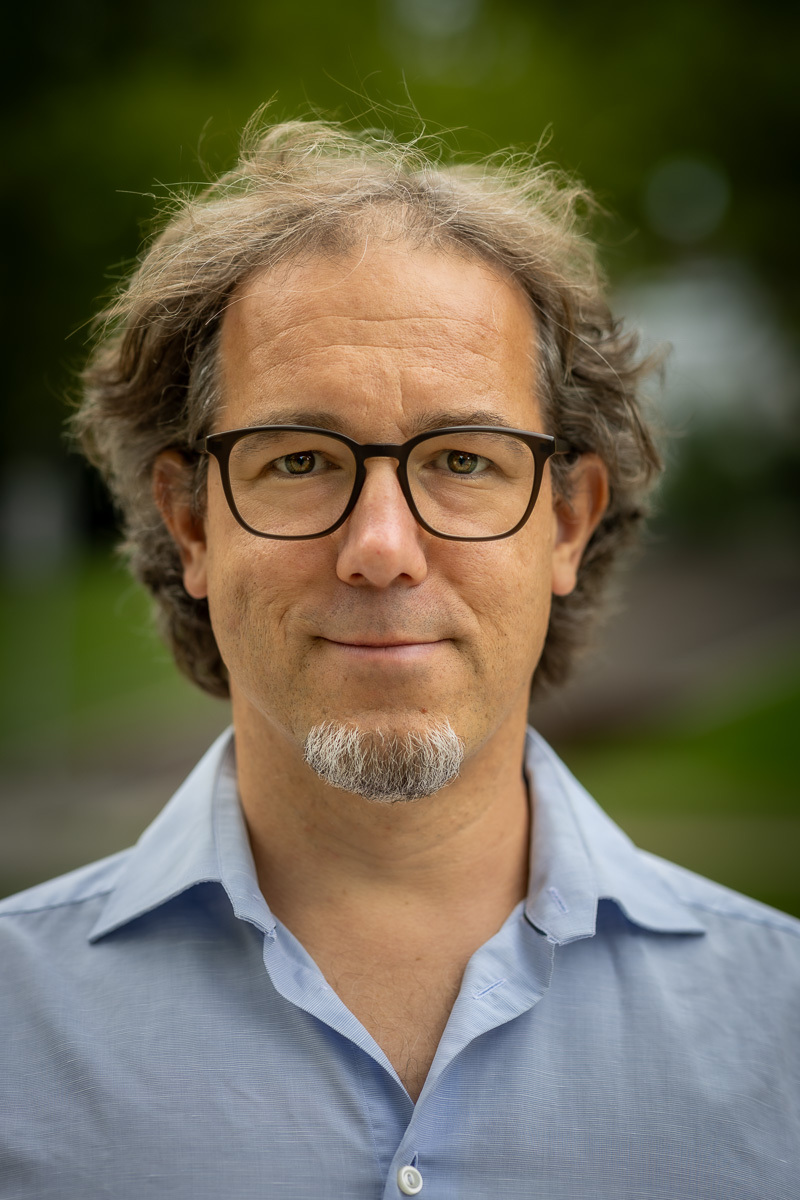}}]{Prof. Dr. Christopher Gies} is leading the quantum theory group at the Carl von Ossietzky University Oldenburg, Germany. His main research direction is the theoretical modelling of nanomaterials at the intersection of quantum optics and quantum information technologies, specializing in many-body methods that enable understanding and the prediction of novel physical phenomena on the basis of the underlying microscopic processes and correlation effects. He has studied and worked at universities in Berlin, Otago (New Zealand), and Bremen, obtaining his MSc degree on ultracold Bose gases, his PhD on quantum dots in microcavities, and his habilitation on semiconductor sources for coherent and quantum light.
\end{IEEEbiography}

\appendix

\section{Alternative Context Encoding Methods And Results}
\label{app:alternative_encodings}
\subsection{Alternative Context Encoding Methods}

Section~\ref{sec:8} detailed the exponential decay sinusoidal encoding method that demonstrated optimal performance in experiments among all encoding strategies we tested. In this section, we discuss the results of the alternative strategies. Each method offers distinct characteristics in how it encodes word identity, positional information, and inter-word relationships. This appendix presents these alternative approaches, their motivations, and comparative experimental results.

All encoding methods share the common framework of transforming a context window of $n$ words into a matrix $\mathbf{C} \in \mathbb{R}^{n \times n}$ (or vector subsequently reshaped to matrix form), which is then reshaped to $\mathbf{\tilde{C}} \in \mathbb{R}^{2m \times \lceil n^2/2m \rceil}$ and padded with zeros as needed for circuit encoding. The fundamental vocabulary-based angle $\theta_i = \text{idx}_i \cdot \frac{2\pi}{|V|}$ appears across all methods, providing a consistent basis for word identity encoding.

\subsubsection{Index-Based Diagonal Method}

This method introduces asymmetric treatment of diagonal versus off-diagonal elements to emphasize self-relationships while maintaining decay-based encoding for word interactions:

\begin{equation}
c_{ij} = 
\begin{cases}
\log(1 + \text{idx}_i), & \text{if } i = j \\
e^{-\alpha |i-j|} \sin(\omega \theta_i) + \theta_i, & \text{if } i \neq j
\end{cases}
\end{equation}

The logarithmic scaling of diagonal elements $\log(1 + \text{idx}_i)$ provides stronger representation for words with higher vocabulary indices. This design assumes that rarer words (typically assigned higher indices in sorted vocabularies) require more prominent self-representation to prevent being overshadowed by more common words. The off-diagonal elements retain exponential distance decay but use only a single sinusoidal term rather than the product $\sin(\omega \theta_i) \cos(\omega \theta_j)$ employed in the exponential decay method.

\subsubsection{Positional Phase Shift Method}

This approach emphasizes absolute position encoding alongside word identity through position-dependent phase modulation:

\begin{equation}
c_{ij} = e^{-\alpha |i-j|} \sin(\omega i + \delta \theta_i) + \theta_i
\end{equation}

where $\delta > 0$ controls the phase shift scaling. The term $\omega i$ introduces a position-dependent base phase that varies linearly with position in the context window, while $\delta \theta_i$ adds word-specific phase modulation. This combination enables the model to distinguish identical words appearing at different positions—a capability that may prove valuable for tasks requiring fine-grained word order sensitivity beyond simple proximity relationships.

\subsubsection{Hash-Based Modulation Method}

To enhance representational uniqueness and reduce potential collisions in the encoding space, this method incorporates a hash function:

\begin{equation}
c_{ij} = e^{-\alpha |i-j|} \sin(\omega i + h_i) + \theta_i
\end{equation}

where $h_i = (\text{idx}_i \cdot p) \bmod N$ with $p$ being a prime number and $N$ the hash space size. The hash function disperses similar vocabulary indices across the encoding space, potentially improving discrimination when word indices are clustered. The use of prime numbers helps ensure good distributional properties, reducing the likelihood that structurally similar words receive near-identical encodings.

\subsubsection{Positional Angular Shift Vector Method}

Unlike the previous matrix-based methods, this approach begins with a vector representation:

\begin{equation}
v_i = \omega \theta_i
\end{equation}

producing a context vector $\mathbf{v} \in \mathbb{R}^n$ where each element directly encodes the scaled vocabulary angle. This vector is subsequently reshaped into a square matrix for circuit encoding. The method represents the minimal encoding strategy, emphasizing word identity through angular transformation while implicitly encoding position through element ordering. Its simplicity offers computational efficiency but sacrifices the explicit pairwise relationship encoding present in matrix-based methods.

\subsection{Results}

\begin{figure*}[t]
\centering
\begin{subfigure}[t]{0.46\textwidth}
\centering
\includegraphics[width=\linewidth]{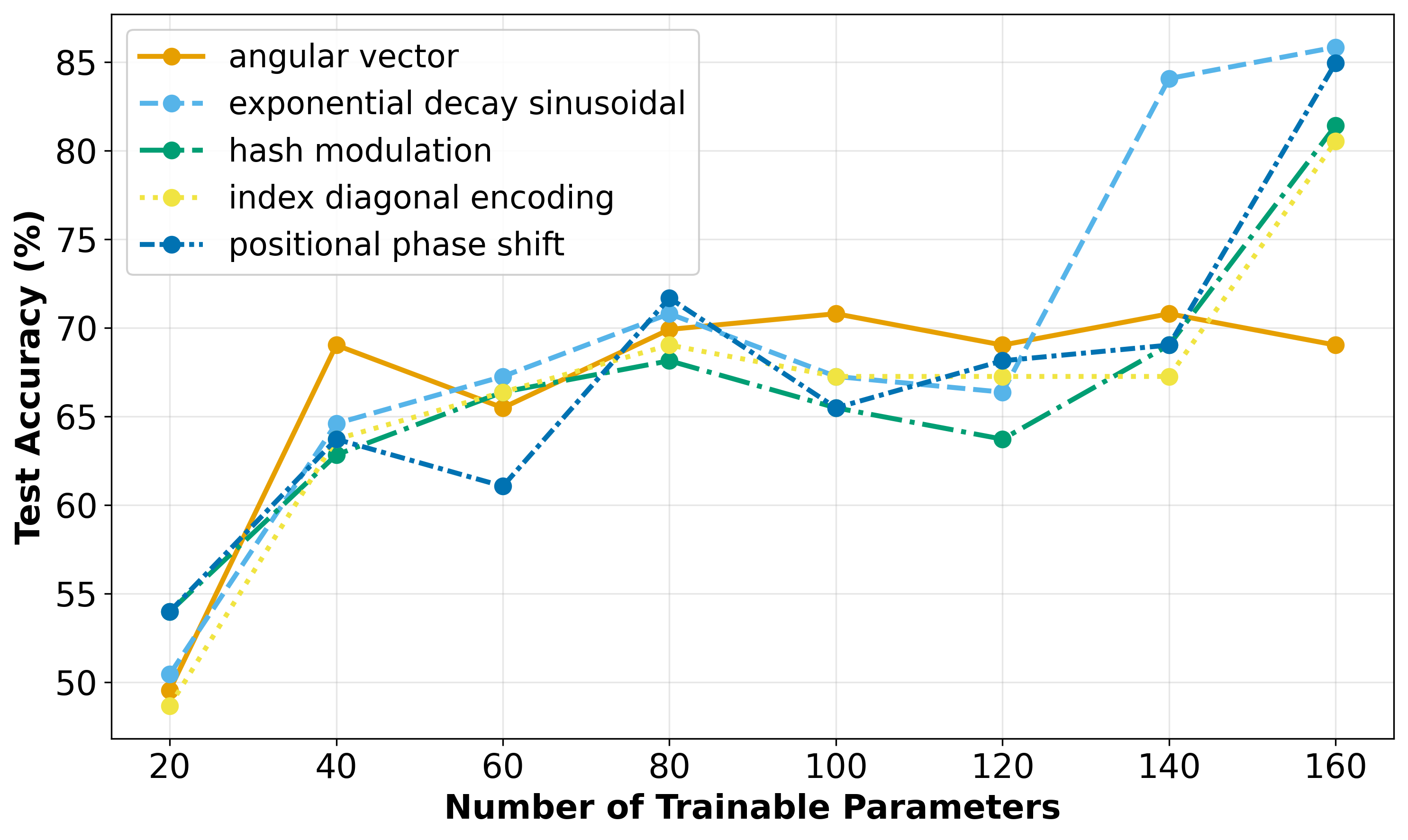}
\caption{Trainable parameter count vs. Accuracy using English dataset}
\label{fig:params_count_eng}
\end{subfigure}
\hfill
\begin{subfigure}[t]{0.46\textwidth}
\centering
\includegraphics[width=\linewidth]{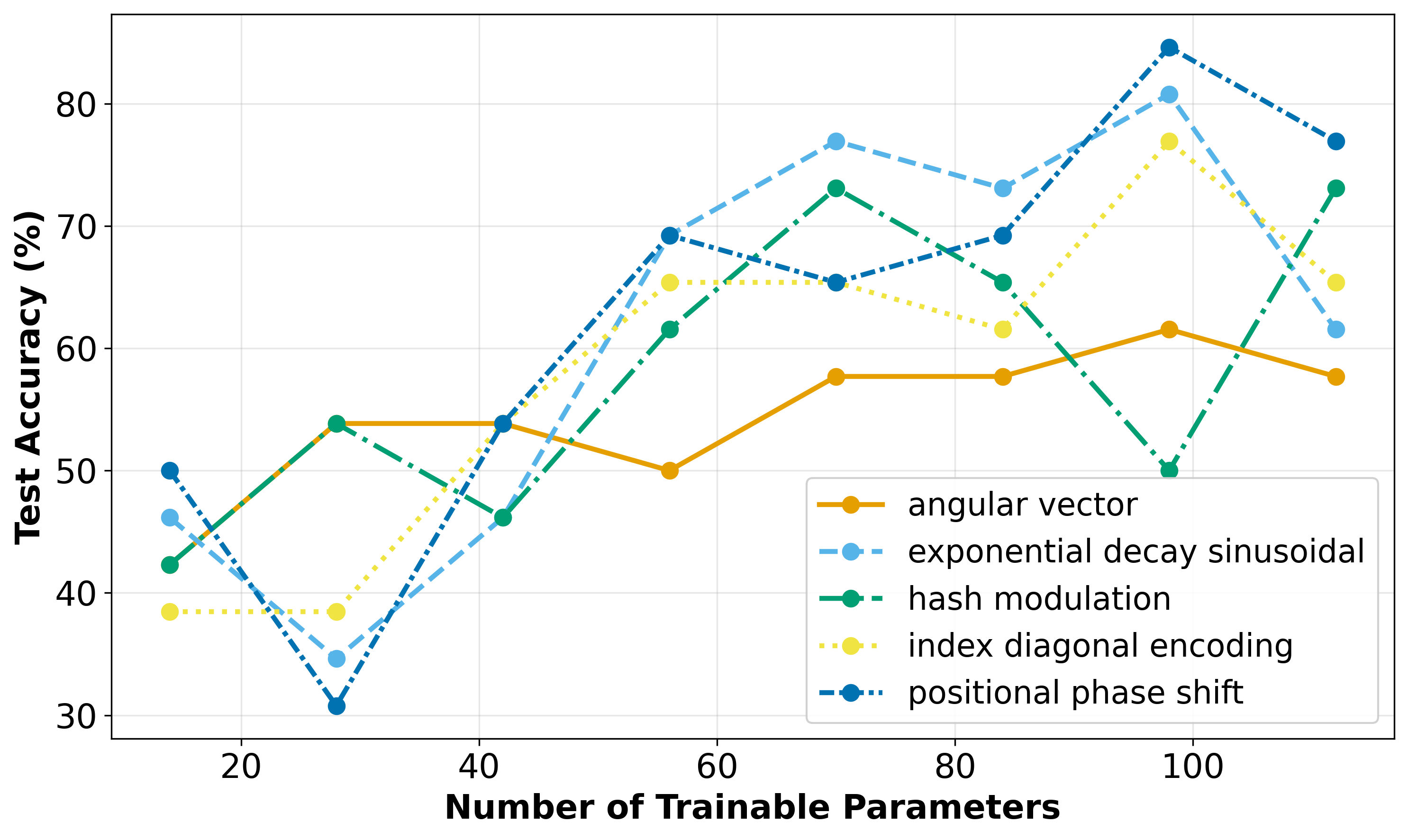}
\caption{Trainable parameter count vs. Accuracy using Fulani dataset}
\label{fig:params_count_ful}
\end{subfigure}
\caption{Accuracy versus trainable parameter count in QCSE. The number of trainable parameters influences performance.}
\label{fig:params_vs_acc_all}
\end{figure*}

\begin{figure*}[t]
\centering
\begin{subfigure}[t]{0.45\textwidth}
\centering
\includegraphics[width=\linewidth]{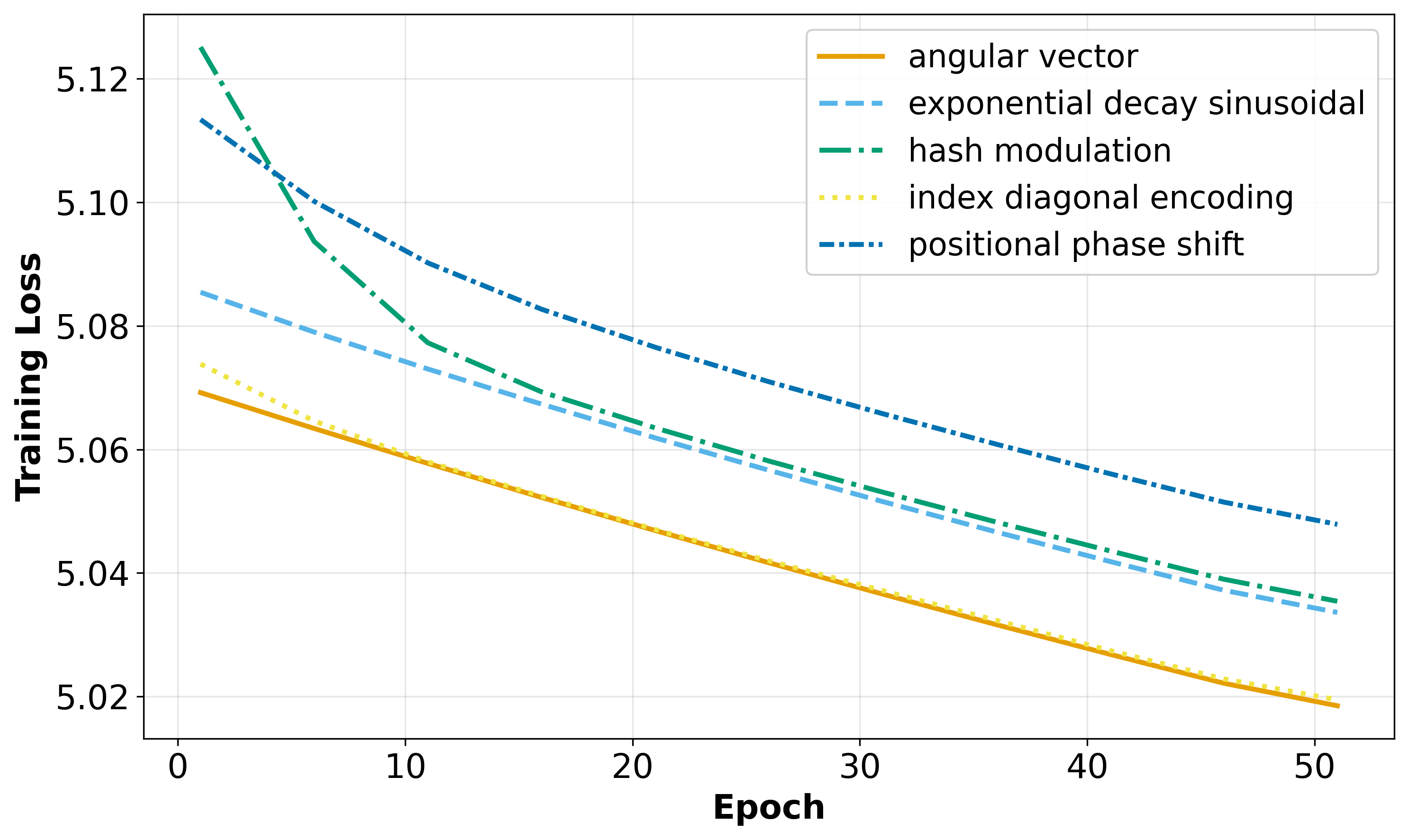}
\caption{QCSE with 1 layer}
\end{subfigure}
\hfill
\begin{subfigure}[t]{0.45\textwidth}
\centering
\includegraphics[width=\linewidth]{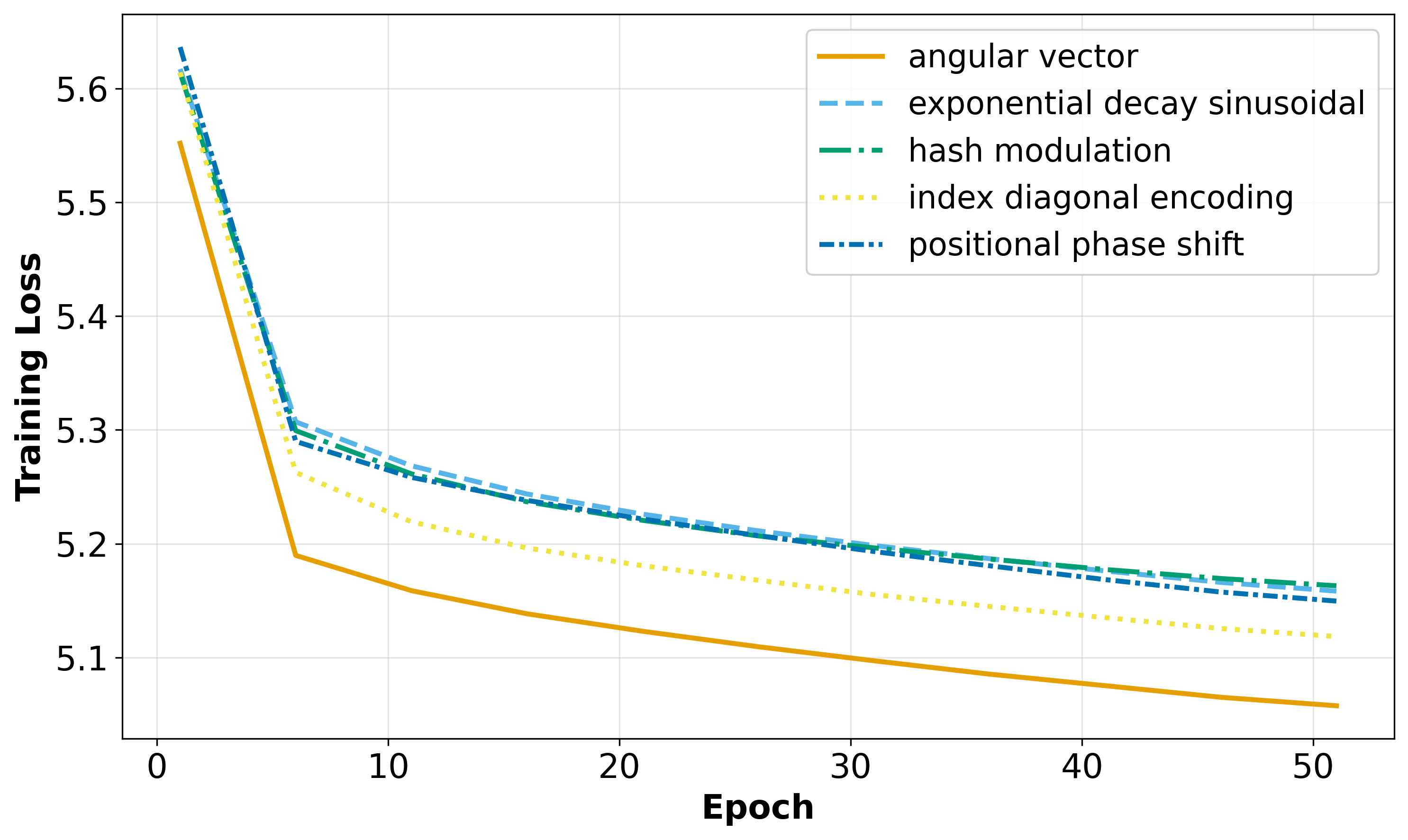}
\caption{QCSE with 2 layers}
\end{subfigure}

\vskip\baselineskip
\begin{subfigure}[t]{0.45\textwidth}
\centering
\includegraphics[width=\linewidth]{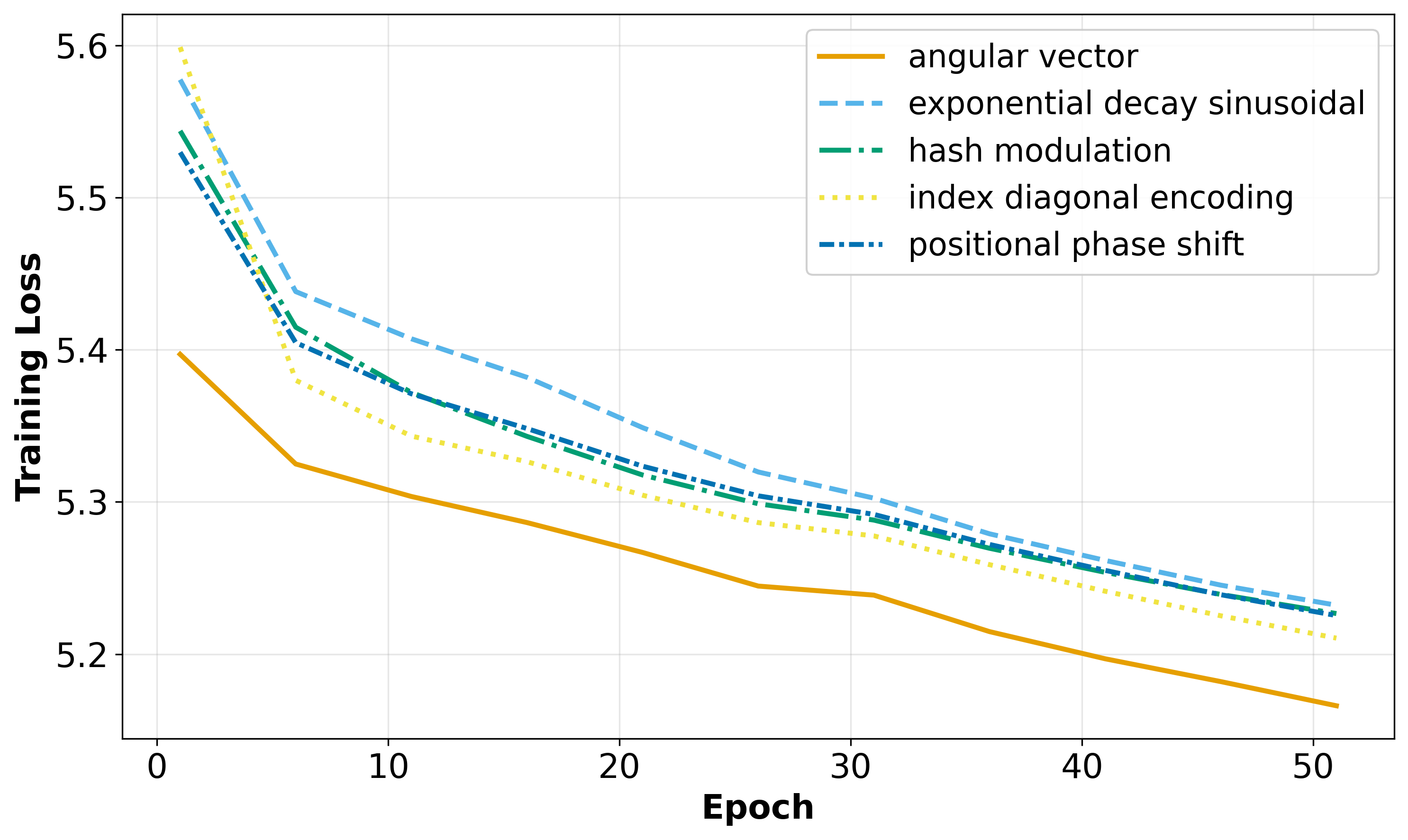}
\caption{QCSE with 3 layers}
\end{subfigure}
\hfill
\begin{subfigure}[t]{0.45\textwidth}
\centering
\includegraphics[width=\linewidth]{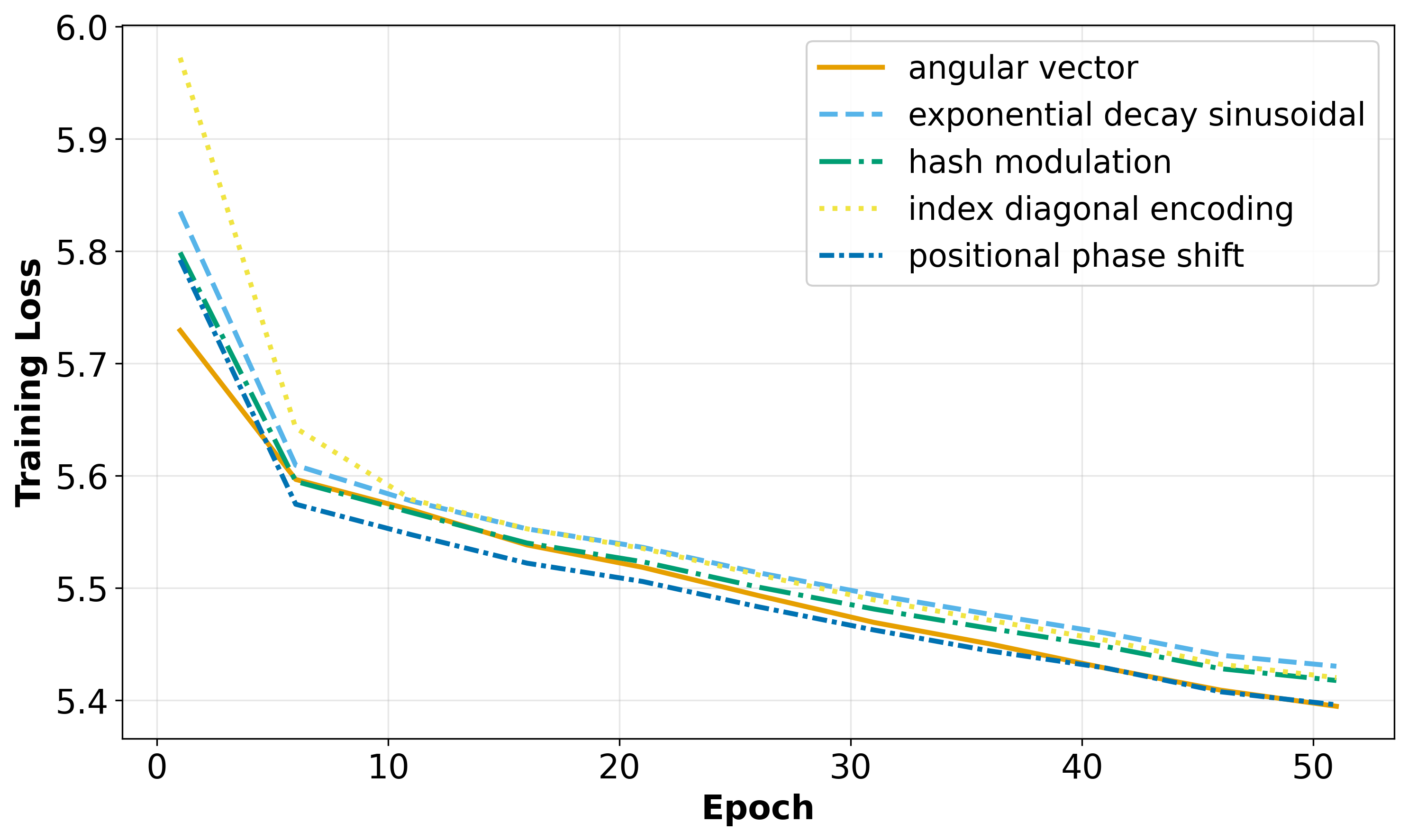}
\caption{QCSE with 4 layers}
\end{subfigure}

\vskip\baselineskip
\begin{subfigure}[t]{0.45\textwidth}
\centering
\includegraphics[width=\linewidth]{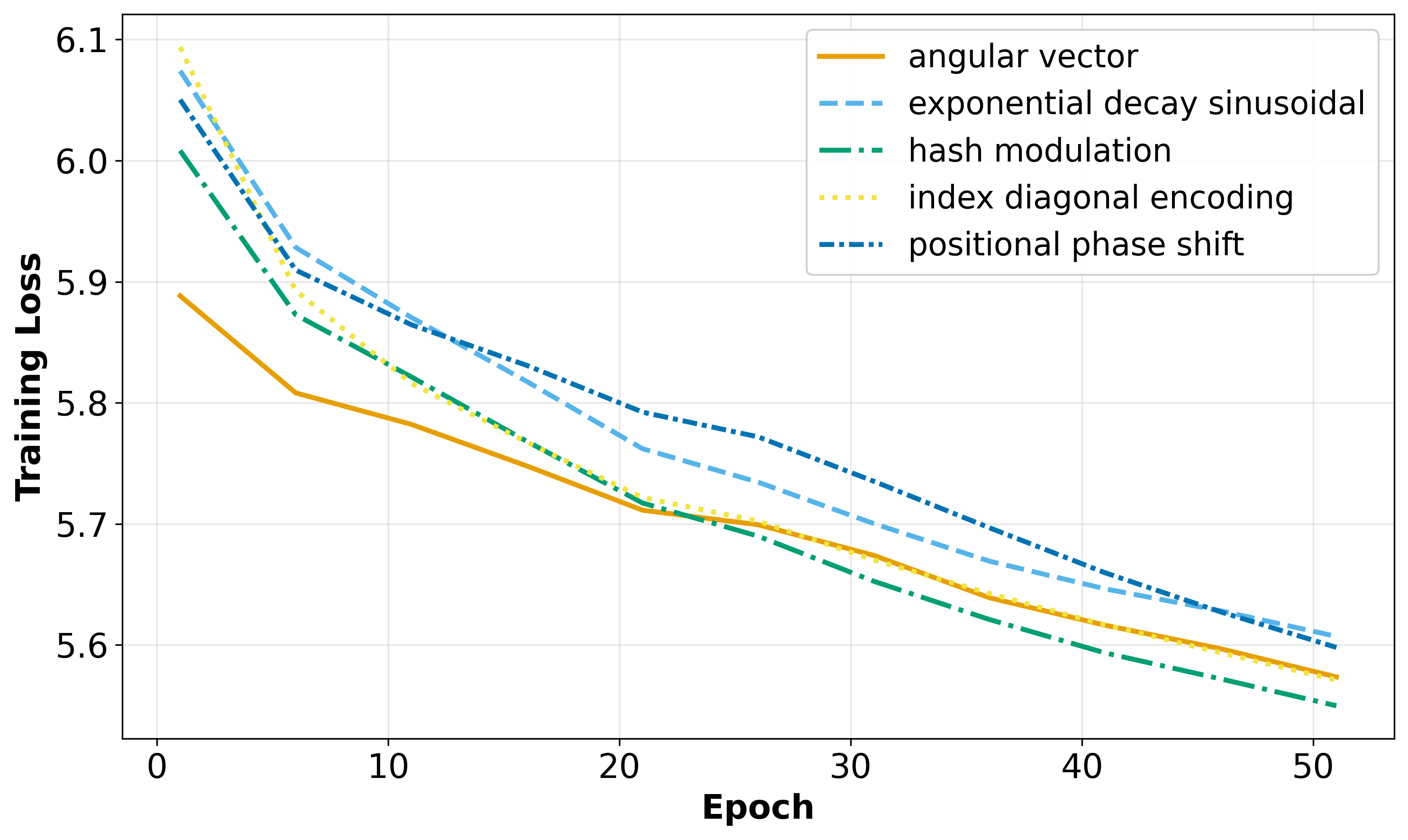}
\caption{QCSE with 5 layers}
\end{subfigure}
\hfill
\begin{subfigure}[t]{0.45\textwidth}
\centering
\includegraphics[width=\linewidth]{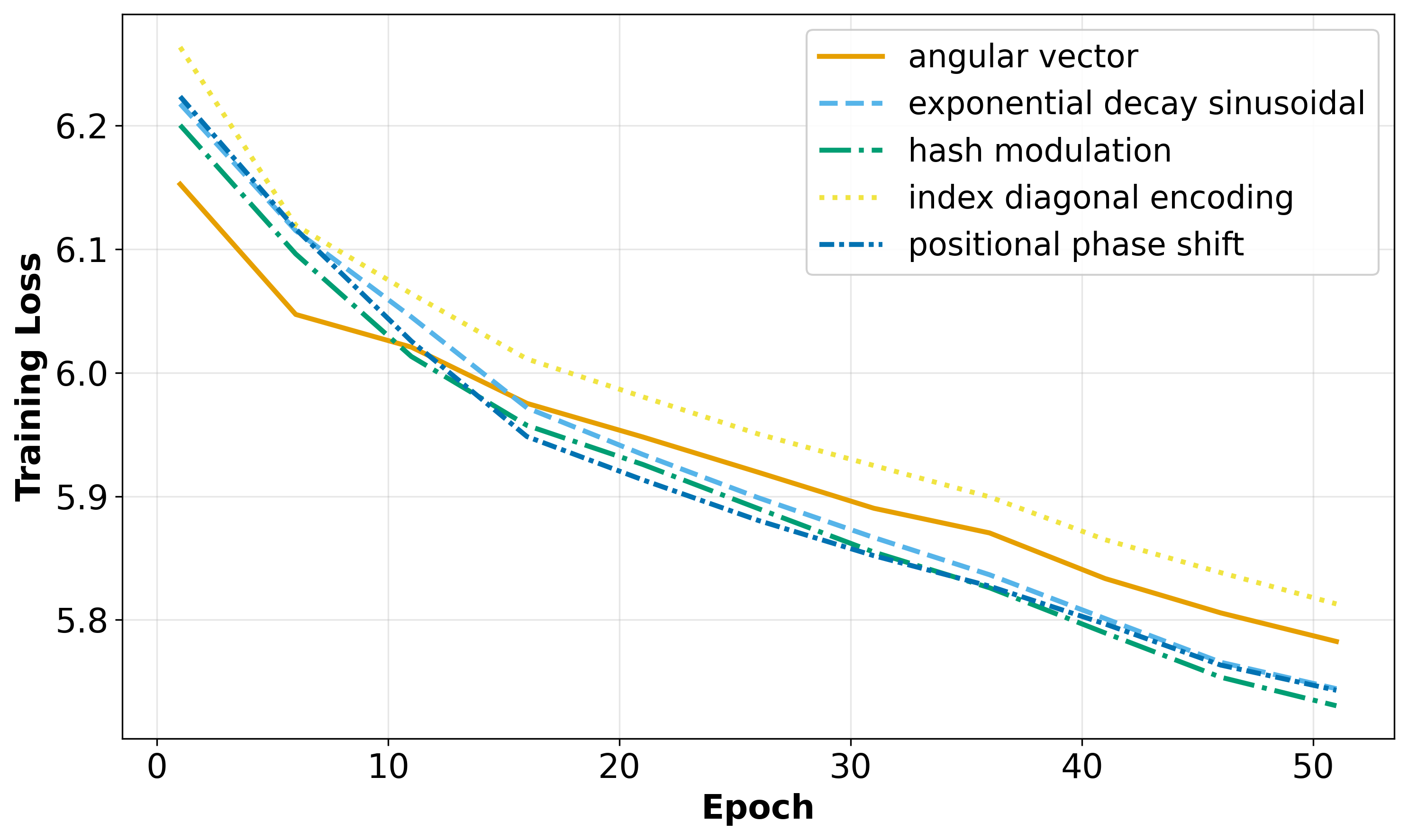}
\caption{QCSE with 6 layers}
\end{subfigure}

\vskip\baselineskip
\begin{subfigure}[t]{0.45\textwidth}
\centering
\includegraphics[width=\linewidth]{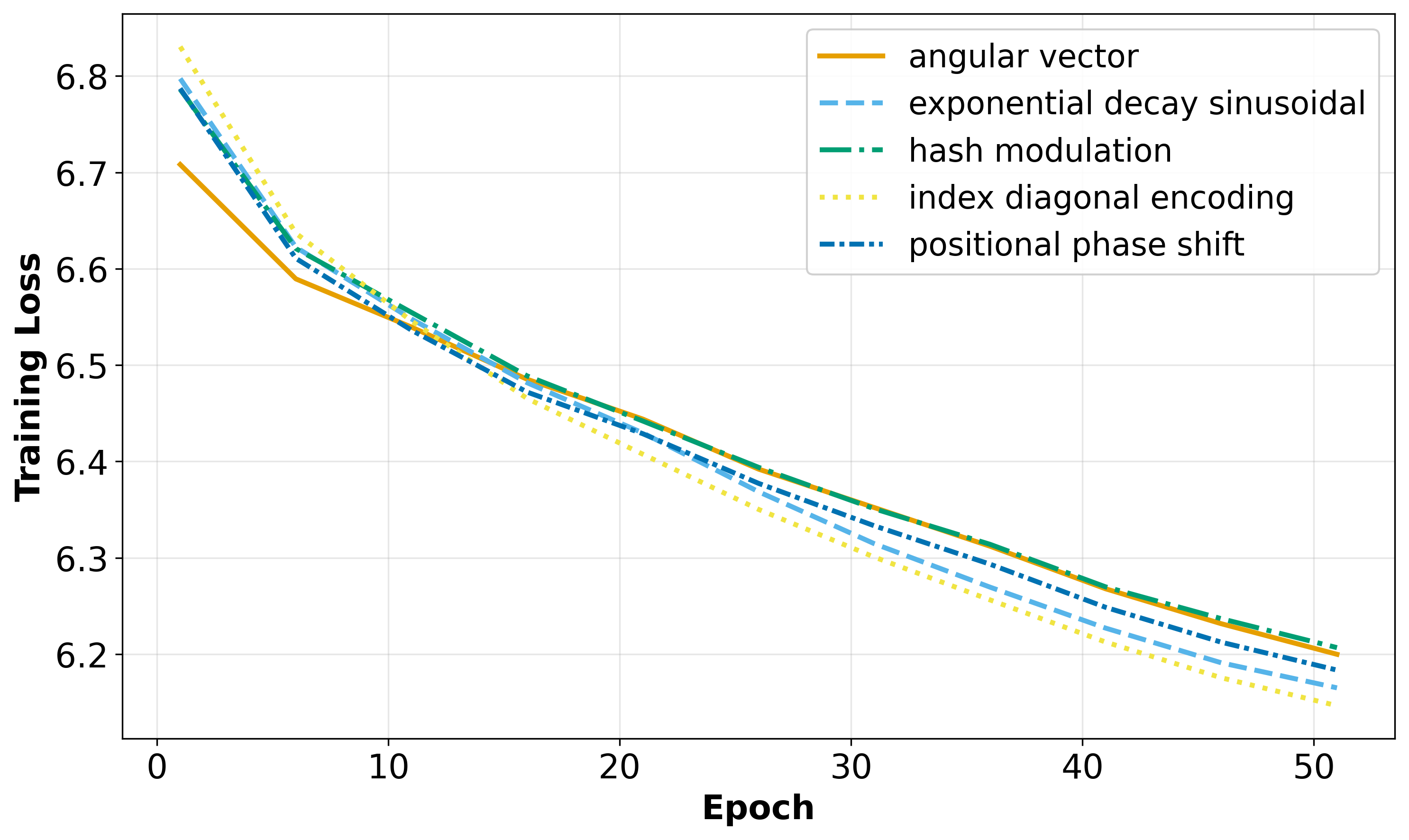}
\caption{QCSE with 7 layers}
\end{subfigure}
\hfill
\begin{subfigure}[t]{0.45\textwidth}
\centering
\includegraphics[width=\linewidth]{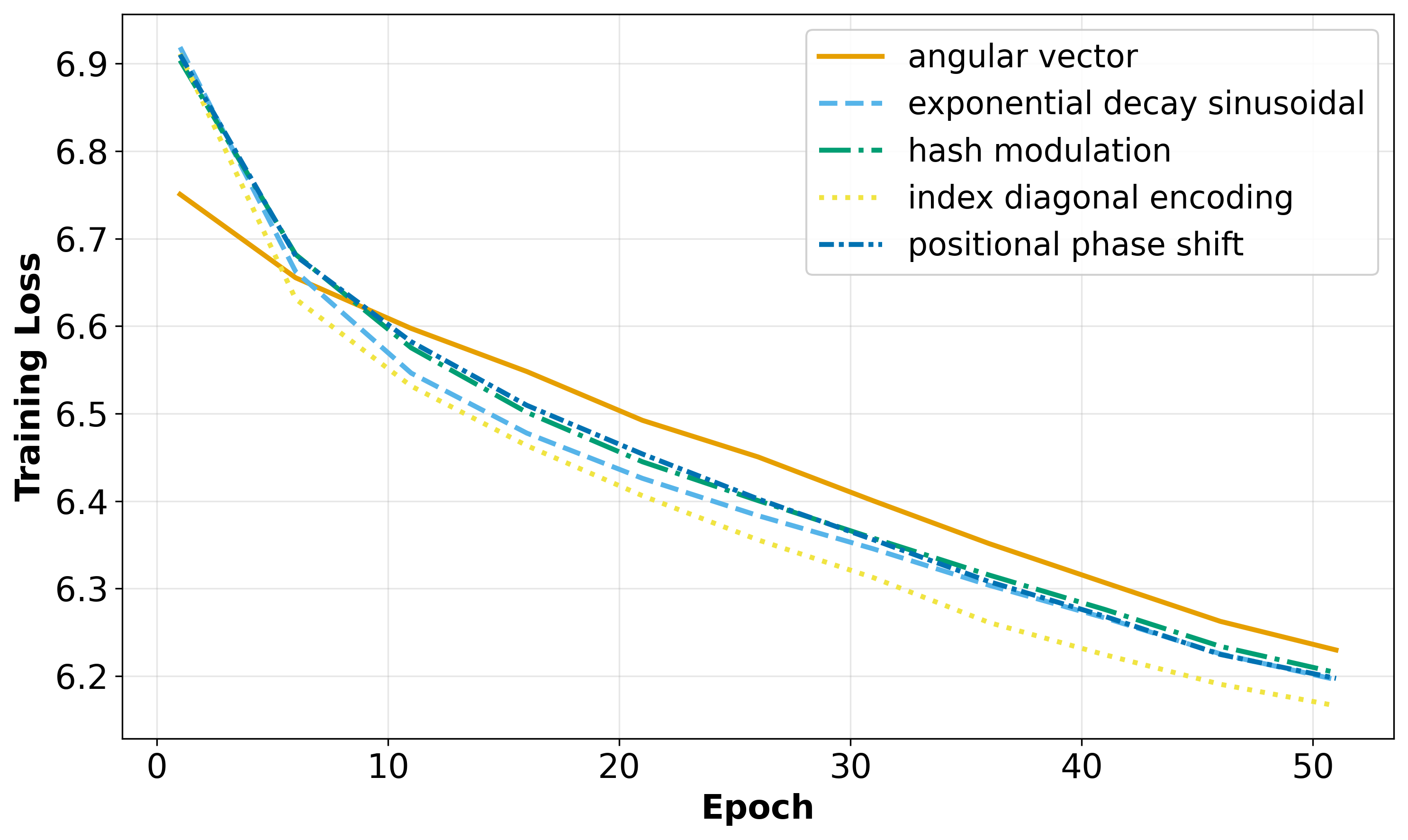}
\caption{QCSE with 8 layers}
\end{subfigure}

\caption{Training loss progression for QCSE models with exponential decay sinusoidal encoding across different depths (1-8 layers) on the English dataset.}
\label{fig:loss_progression_eng}
\end{figure*}

\begin{figure*}[t]
\centering
\begin{subfigure}[t]{0.45\textwidth}
\centering
\includegraphics[width=\linewidth]{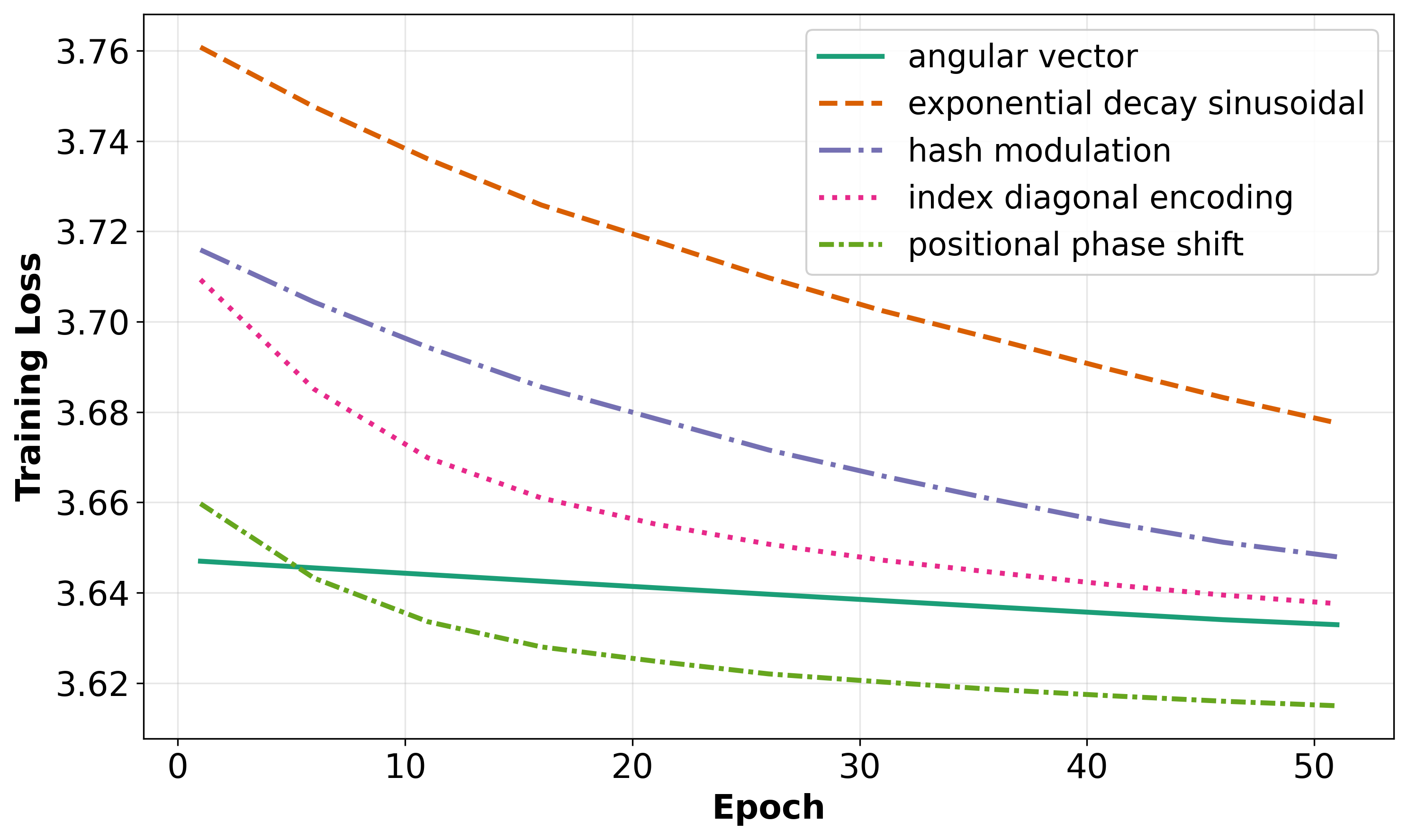}
\caption{QCSE with 1 layer}
\label{fig:loss_ful_L1}
\end{subfigure}
\hfill
\begin{subfigure}[t]{0.45\textwidth}
\centering
\includegraphics[width=\linewidth]{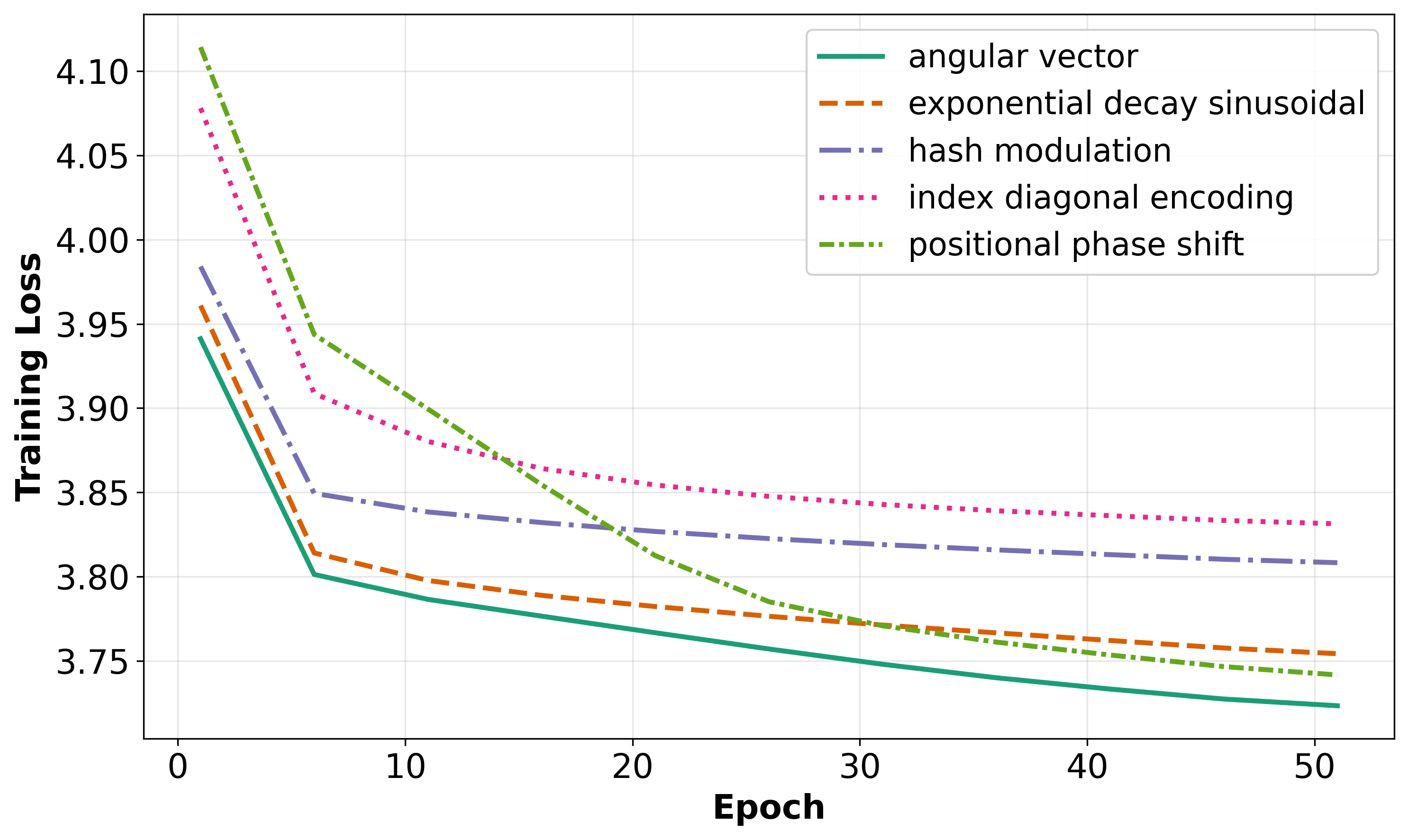}
\caption{QCSE with 2 layers}
\label{fig:loss_ful_L2}
\end{subfigure}

\vskip\baselineskip
\begin{subfigure}[t]{0.45\textwidth}
\centering
\includegraphics[width=\linewidth]{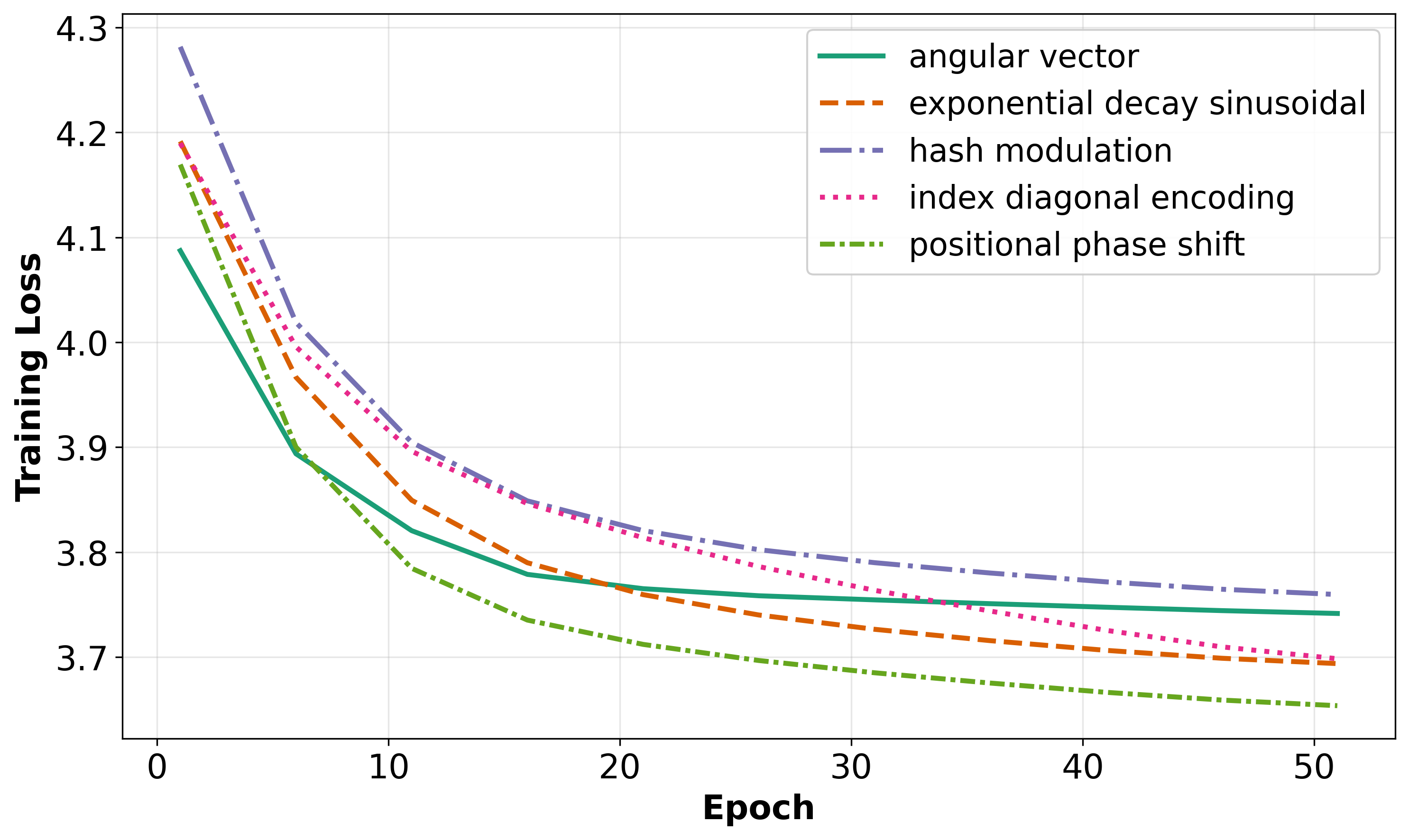}
\caption{QCSE with 3 layers}
\label{fig:loss_ful_L3}
\end{subfigure}
\hfill
\begin{subfigure}[t]{0.45\textwidth}
\centering
\includegraphics[width=\linewidth]{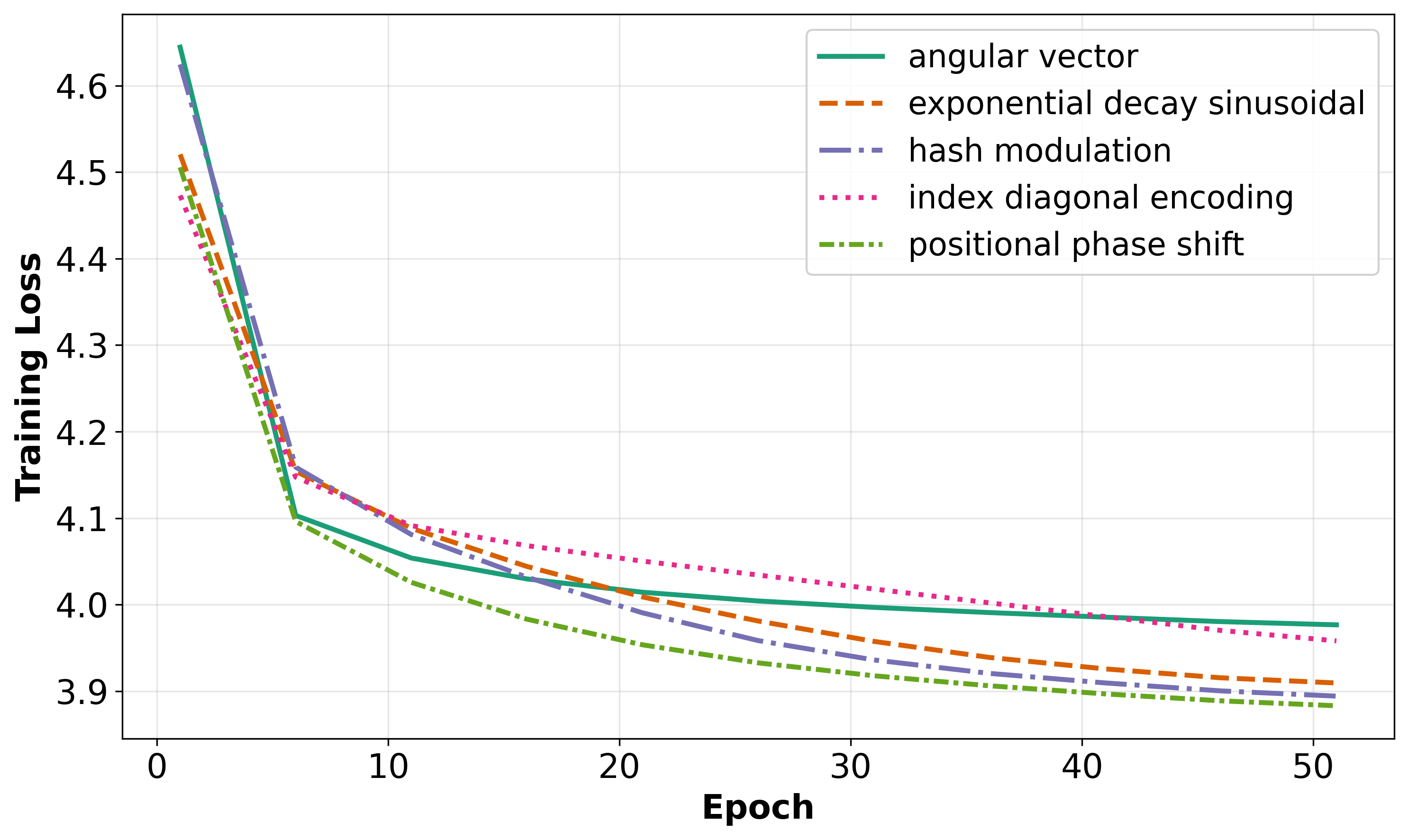}
\caption{QCSE with 4 layers}
\label{fig:loss_ful_L4}
\end{subfigure}

\vskip\baselineskip
\begin{subfigure}[t]{0.45\textwidth}
\centering
\includegraphics[width=\linewidth]{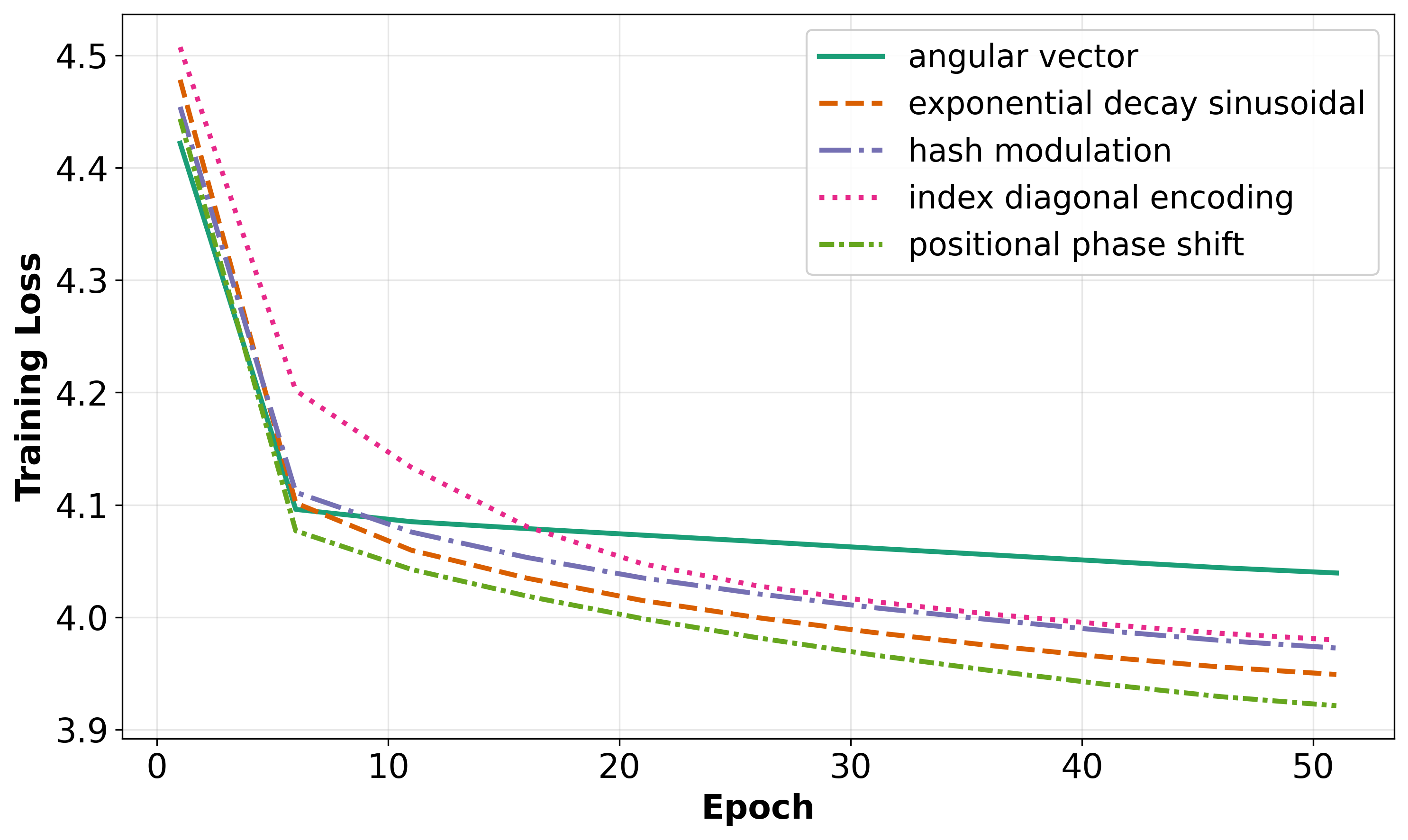}
\caption{QCSE with 5 layers}
\label{fig:loss_ful_L5}
\end{subfigure}
\hfill
\begin{subfigure}[t]{0.45\textwidth}
\centering
\includegraphics[width=\linewidth]{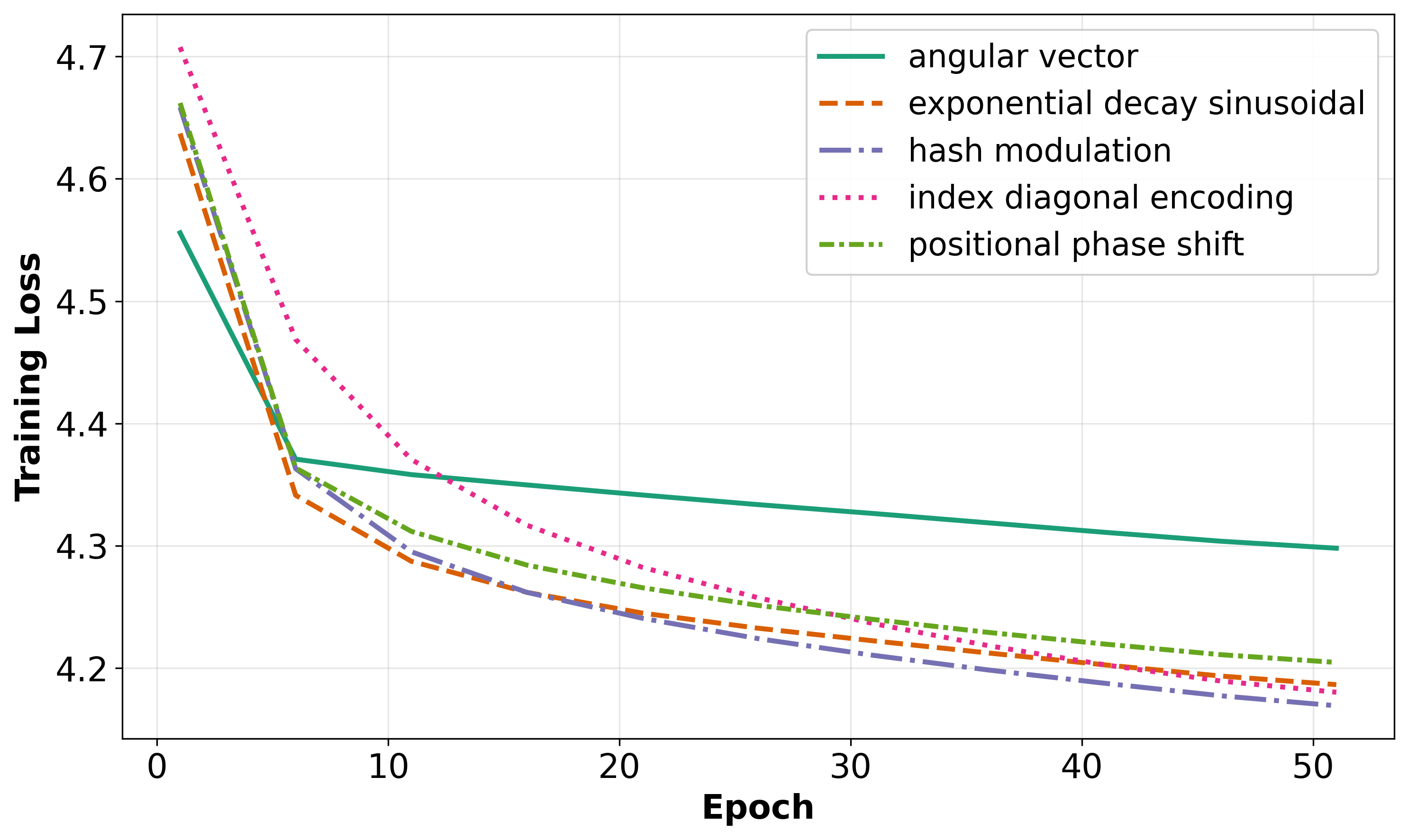}
\caption{QCSE with 6 layers}
\label{fig:loss_ful_L6}
\end{subfigure}

\vskip\baselineskip
\begin{subfigure}[t]{0.45\textwidth}
\centering
\includegraphics[width=\linewidth]{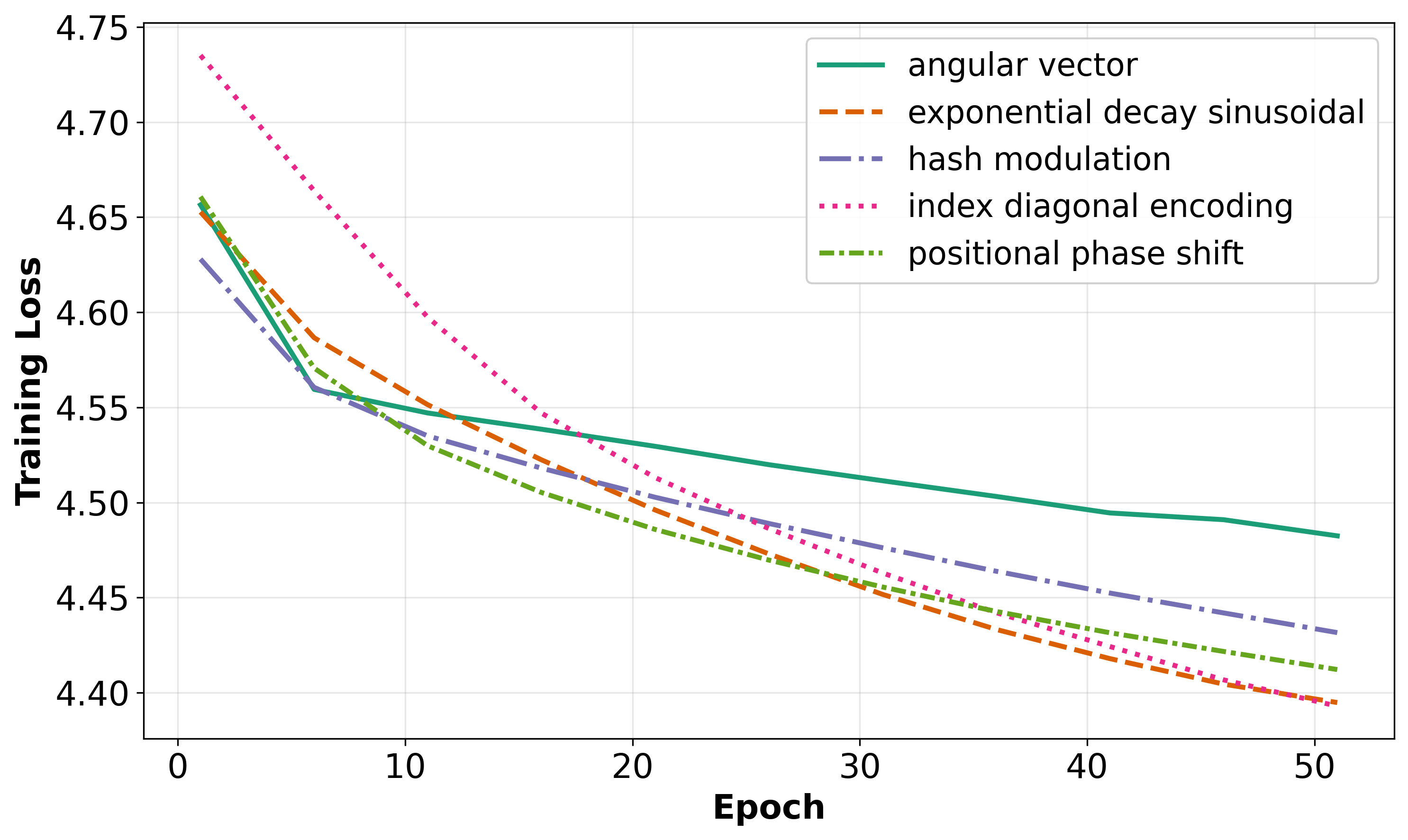}
\caption{QCSE with 7 layers}
\label{fig:loss_ful_L7}
\end{subfigure}
\hfill
\begin{subfigure}[t]{0.45\textwidth}
\centering
\includegraphics[width=\linewidth]{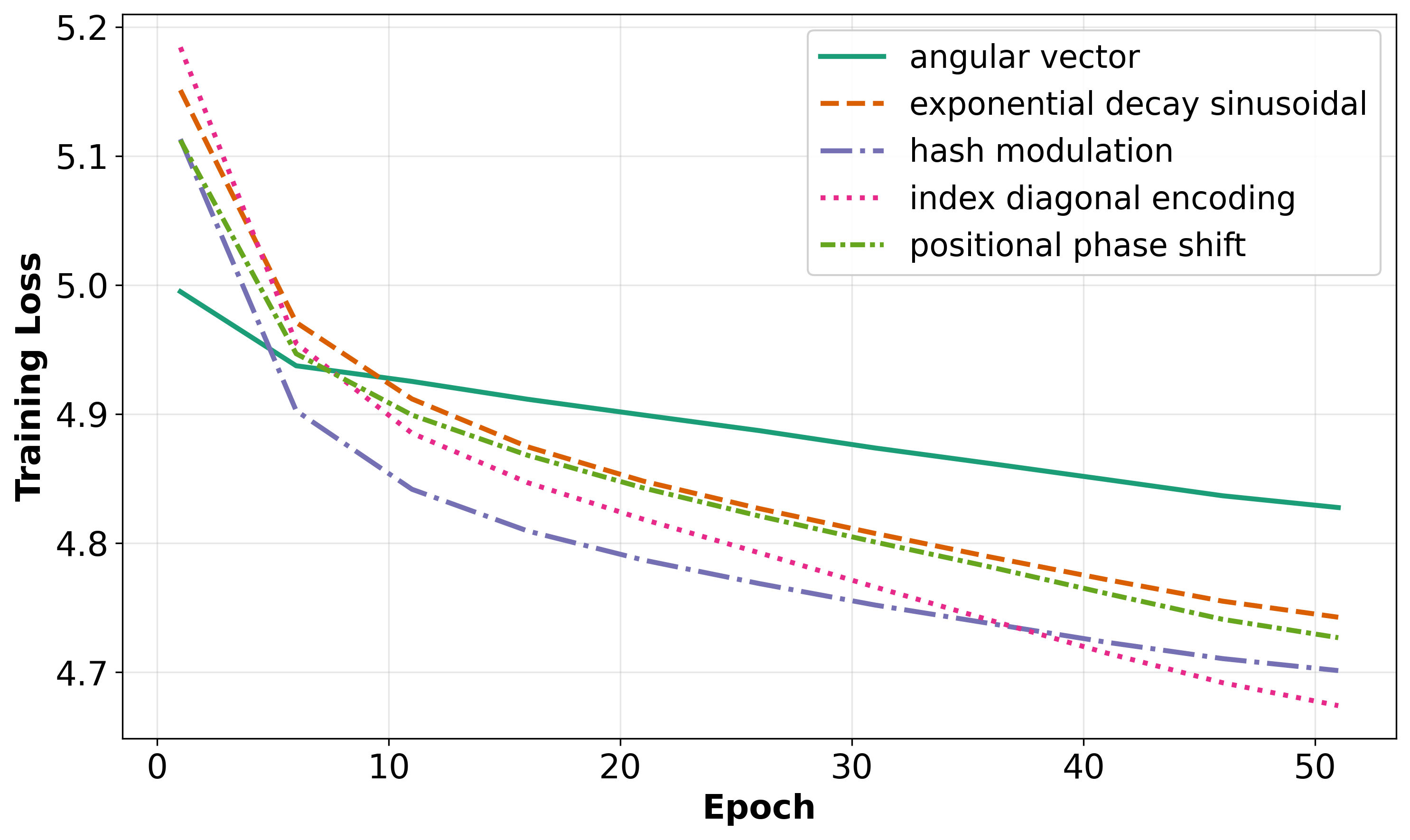}
\caption{QCSE with 8 layers}
\label{fig:loss_ful_L8}
\end{subfigure}

\caption{Training loss plots of QCSE using same number of layers with different context-encoding methods on Fulani dataset.}
\label{fig:loss_comparison_ful}
\end{figure*}

Tables~\ref{tab:alternative_english} and~\ref{tab:alternative_Fulani}, and Figures ~\ref{fig:params_vs_acc_all}, ~\ref{fig:loss_progression_eng} and ~\ref{fig:loss_comparison_ful} present the accuracy results and the training dynamics for all encoding methods across different model depths on both datasets.

\begin{table*}[h]
\centering
\caption{Accuracy results for alternative context encoding methods on English dataset}
\begin{tabular}{lccccc}
\hline
\textbf{Model} & \textbf{Params} & \textbf{Index Diagonal} & \textbf{Phase Shift} & \textbf{Hash Mod.} & \textbf{Angular Vector} \\
\hline
QCSE-1L & 21 & 48.7 & 54.0 & 54.0 & 49.6 \\
QCSE-2L & 42 & 63.7 & 63.7 & 62.9 & 69.1 \\
QCSE-3L & 63 & 66.4 & 61.1 & 66.4 & 65.5 \\
QCSE-4L & 84 & 69.1 & 71.7 & 68.1 & 69.9 \\
QCSE-5L & 105 & 67.3 & 65.5 & 65.5 & 70.8 \\
QCSE-6L & 126 & 67.3 & 68.1 & 63.7 & 69.1 \\
QCSE-7L & 147 & 67.3 & 69.0 & 69.1 & 70.8 \\
QCSE-8L & 168 & 80.5 & 85.0 & 81.4 & 69.1 \\
\hline
\end{tabular}
\label{tab:alternative_english}
\end{table*}

\begin{table*}[h]
\centering
\caption{Accuracy results for alternative context encoding methods on Fulani dataset}
\begin{tabular}{lccccc}
\hline
\textbf{Model} & \textbf{Params} & \textbf{Index Diagonal} & \textbf{Phase Shift} & \textbf{Hash Mod.} & \textbf{Angular Vector} \\
\hline
QCSE-1L & 15 & 38.5 & 50.0 & 42.3 & 42.3 \\
QCSE-2L & 30 & 38.5 & 30.8 & 53.8 & 53.8 \\
QCSE-3L & 45 & 53.8 & 53.8 & 46.2 & 53.8 \\
QCSE-4L & 60 & 65.4 & 69.2 & 61.5 & 50.0 \\
QCSE-5L & 75 & 65.4 & 65.4 & 73.1 & 57.7 \\
QCSE-6L & 90 & 61.5 & 69.4 & 65.4 & 57.7 \\
QCSE-7L & 105 & 76.9 & 84.6 & 50.0 & 61.5 \\
QCSE-8L & 120 & 65.4 & 76.9 & 73.1 & 57.7 \\
\hline
\end{tabular}
\label{tab:alternative_Fulani}
\end{table*}

The alternative encoding methods exhibit variable and comparable performance compared to the exponential decay sinusoidal approach. On the English dataset, peak accuracies range from 69.1-85.0\% (compared to 85.8\% for exponential decay), with optimal depths varying by method: Index-Based Diagonal Method peaks at 8 layers (80.5\%), Positional Phase Shift Method at 8 layers (85.0\%), Hash-Based Modulation Method at 8 layers (81.4\%), and Positional Angular Shift Vector Method plateaus at 5-7 layers (70.8\%). This variability means that the encoding strategy influences the performance of the quantum circuit and that larger and more diverse datasets are necessary for reliably discriminating between encoding strategies in quantum embedding models.

The Positional Angular Shift Vector Method demonstrates limited performance, maintaining 69-71\% accuracy across 2-8 layers on English data. This reflects its simpler encoding structure—without explicit pairwise relationship terms, the model reaches its representational limit early and cannot benefit from additional circuit depth. In contrast, Index Diagonal and Hash Modulation methods show continued improvement through 8 layers, suggesting that their encodings provide sufficient representational to exploit deeper circuits, though outperformed by the exponential decay method.

On the smaller Fulani dataset, some alternative methods exhibit instability. The Hash-Based Modulation method ranges from 42.3\% (1L) to 73.1\% (5L and 8L), with a collapse at 7 layers (50.0\%). The Positional Phase Shift method achieves the highest peak (84.6\% at 7L) but shows variation across depths (30.8\% at 2L). This instability contrasts with the exponential decay method progression, suggesting that the product term $\sin(\omega \theta_i) \cos(\omega \theta_j)$ in Equation~\ref{eq:exp_decay} provides robust optimization properties. The Index-Based Diagonal encoding using a self-relationships (via $\log(1 + \text{idx}_i)$ diagonal terms) achieves comparable performance (80.5\% at 8L on English) but does not outperform the exponential decay approach. The Positional Phase Shift method employing an emphasis on absolute position (through $\omega i$) yields the second-best overall result (85.0\%), nearly matching exponential decay, suggesting that position encoding complementary to relative distance may merit further investigation. The Hash-Based Modulation mechanism shows promise (81.4\%) but introduces additional hyperparameter sensitivity (prime $p$, hash space $N$) not present in exponential decay.

\label{app:loss_analysis}


Figures~\ref{fig:loss_progression_eng} and~\ref{fig:loss_comparison_ful} show depth-dependent training dynamics across encoding methods. Initial loss increases with circuit depth (5.0 for 1 layer to 6.2–6.9 for 7–8 layers), reflecting the higher-dimensional optimization landscape of deeper circuits. Three convergence regimes emerge: shallow models (1–2 layers) exhibit minimal loss reduction and early plateau, leading to poor accuracy; medium-depth models (3–6 layers) show steady descent and moderate performance; and deep models (7–8 layers) display more loss reduction, achieving the highest accuracies. Overall, effective learning in quantum embedding models is shown across all the context encoding methods.

\end{document}